\documentclass[10pt,twocolumn,letterpaper]{article}

\usepackage[pagenumbers]{cvpr} 
\usepackage{bm}
\definecolor{cvprblue}{rgb}{0.21,0.49,0.74}
\usepackage[pagebackref,breaklinks,colorlinks,allcolors=cvprblue]{hyperref}

\def\confName{CVPR}
\def\confYear{2027}

\title{ FairTalking-Bench: A Large-Scale Demographic-Balanced Talking-Head Video Benchmark for Deepfake Detection}

\author{
    Jiao Wu$^{1}$,\quad Yaning Zhang$^{2}$,\quad ZhiHao Lao$^{1}$,\quad Bizhu Wu$^{3} $
    Lintian Feng$^{1}$,\quad Xiaoqin Wang$^{1}$,\quad Linlin Shen$^{1}$\\[0.5em]
    $^{1}$Shenzhen University $^{2}$Qilu University of Technology $^{3}$The University of Nottingham Ningbo China
}

\begin{document}
\maketitle
\begin{abstract}
 Recent advances in talking-head generation have increased the realism and diversity of Deepfakes. However, existing Deepfake video benchmarks predominantly focus on GAN-based face swapping and do not consider demographic balance and modern generation paradigms. To address this gap, we introduce FairTalking-Bench, a large-scale and demographically balanced talking-head video benchmark for fair Deepfake detection, which comprises 26,676 videos encompassing eight advanced generation algorithms (\emph{e.g.}, Diffusion, Flow-matching, AE/GAN, and 3DMM), with demographic balance across three attributes in 48 subgroups—gender (2 categories), age (6 groups), and race (4 groups). Based on FairTalking-Bench, we design four protocols to evaluate the performance of deepfake detectors under zero-shot robustness, demographic fairness, cross-generator generalization, and identity-consistent detection settings. Extensive experimental results suggest that generalization-oriented orthogonal representation decomposition may improve zero-shot transfer. Demographic balance improves detection performance and reduces most group disparities. Flow-based generators challenge existing detectors but offer promising training distributions for improved cross-generator generalization. Pretrained representations promote cross-generator generalization, whereas generator-specific artifacts hinder it. Finally, identity-paired evaluation reveals identity-related shortcuts, motivating the disentanglement of identity information from forgery-related features. The dataset and code will be made publicly available upon acceptance.
\end{abstract}

\vspace{-20pt}

\section{Introduction}
\label{sec:intro}

The rapid evolution of Generative Artificial Intelligence (GenAI) has fundamentally reshaped the landscape of digital media. While early deepfake technologies are primarily confined to GAN-based face swapping, the field has recently witnessed a paradigm shift toward audio-driven talking-head generation. Talking-head generation aims to synthesize photorealistic videos with intricate lip synchronization and natural head poses from a single reference image and an audio clip, via advanced architectures such as Latent Diffusion Models (LDMs)~\cite{ji2025sonic}, Flow-matching~\cite{ki2025float}, and 3D Morphable Models (3DMM)~\cite{zhang2023sadtalker}. While these methods achieve increasingly realistic synthesis with natural lip synchronization and coherent head movement, they also lower the barrier for malicious exploitation, posing unprecedented risks to information security and societal trust~\cite{wired2024storm1516}. Therefore, developing a large-scale talking-head benchmark is crucial for advancing the development of deepfake detection.

However, there are some limitations for existing deepfake benchmarks, including the talking-head benchmark. \textbf{ Obsolescence: }As Table~\ref{tab:deepfake_dataset_demographics} shows, conventional benchmarks like FaceForensics++~\cite{rossler2019faceforensics++} (FF++) mainly contain face forgery videos generated by GAN, which fail to include advanced diffusion or Flow-matching forgery distributions.~\textbf{ Demographic Bias:} In Table~\ref{tab:deepfake_dataset_demographics}, most current datasets lack rigorous control over demographic attributes (\emph{e.g.}, race, gender, and age). For instance, although DF-Platter~\cite{Narayan_2023_CVPR}  includes demographic labels, its coverage is limited to Indian subjects only; other benchmarks such as TalkingHeadBench~\cite{Xiong_2026_WACV} provide no demographic annotations at all. Such limited demographic diversity hinders fair and comprehensive evaluation across different population groups.~\textbf {Lack of Systematic Evaluation:} Current benchmarks~\cite{Xiong_2026_WACV} hardly evaluate the generalization of detectors to unseen State-of-the-Art (SOTA) generation paradigms (\emph{e.g.}, Flow-matching),  the robustness to challenging adversarial scenarios, such as identity-consistent forgeries, and the fairness to demographic attribute settings.

\begin{table*}[t]
  \centering
    \caption{\textbf{Comparison of deepfake video benchmarks.} FairTalking-Bench covers a broader range of talking-head forgery methods with balanced demographics and includes an ID-paired subset. \textbf{DM}: Diffusion Models, \textbf{FLOW}: Flow-matching model, \textbf{AE}: Autoencoder, and \textbf{HC} : hand-crafted methods.}
  \label{tab:deepfake_dataset_demographics}
  \vspace{-5pt}
  \resizebox{\linewidth}{!}{%
  \renewcommand{\arraystretch}{0.95}
  \scriptsize
  \setlength{\tabcolsep}{4pt}
  \begin{tabular}{l r c c c c c c r c c c c c}
  \toprule
    \textbf{Dataset} & 
    \textbf{Year} & 
    \textbf{Count} & 
    \multicolumn{6}{c}{\textbf{Deepfake Model}} & 
    \multirow{2}{*}{\shortstack{\textbf{ID-}\\\textbf{Paired}}} & 
    \multirow{2}{*}{\textbf{Balanced}} & 
    \multicolumn{3}{c}{\textbf{Demographics}} \\
    
    \cmidrule(lr){3-3} \cmidrule(lr){4-9} \cmidrule(lr){12-14}
    
     & & Real/Fake & GAN & DM & FLOW & AE & HC & \textbf{No.} & & & Race & Gender & Age \\
    \midrule
    UADFV~\cite{UADFV}                               & 2018 & 49/49                    & $\times$      & $\times$      & $\times$      & $\times$      & $\checkmark$  & 1  & $\times$      & $\times$      & $\times$      & $\times$      & $\times$      \\
    DeepfakeTIMIT~\cite{DeepfakeTIMIT}               & 2018 & 320/640                  & $\checkmark$  & $\times$      & $\times$      & $\times$      & $\times$      & 2  & $\times$      & $\times$      & $\times$      & $\times$      & $\times$      \\
    FaceForensics++~\cite{rossler2019faceforensics++} & 2019 & 1{,}000/4{,}000       & $\times$      & $\times$      & $\times$      & $\checkmark$  & $\checkmark$  & 4  & $\times$      & $\times$      & $\times$      & $\times$      & $\times$      \\
    DFD~\cite{rossler2019faceforensics++}               & 2019 & 363/3{,}068              & --            & --            & --            & --            & --            & 5  & $\times$      & $\times$      & $\times$      & $\times$      & $\times$      \\
    Celeb-DF-v1~\cite{li2020celeb}                   & 2020 & 590/795                  & $\times$      & $\times$      & $\times$      & $\checkmark$  & $\times$      & 1  & $\times$      & $\times$      & $\times$      & $\times$      & $\times$      \\
    DFDC~\cite{dolhansky2020deepfake}                & 2020 & 23{,}654/104{,}500       & $\checkmark$  & $\times$      & $\times$      & $\checkmark$  & $\checkmark$  & 8  & $\times$      & $\times$      & $\times$      & $\times$      & $\times$      \\
    DeeperForensics-1.0~\cite{jiang2020deeperforensics} & 2020 & 50{,}000/10{,}000   & $\checkmark$  & $\times$      & $\times$      & $\times$      & $\times$      & 1  & $\times$      & $\times$      & $\times$      & $\times$      & $\times$      \\
    WildDeepfake~\cite{zi2020wilddeepfake}           & 2021 & 3{,}805/3{,}509          & --            & --            & --            & --            & --            & -- & $\times$      & $\times$      & $\times$      & $\times$      & $\times$      \\
    ForgeryNet~\cite{he2021forgerynet}               & 2021 & 99{,}630/121{,}617       & $\checkmark$  & $\times$      & $\times$      & $\checkmark$  & $\checkmark$  & 8  & $\times$      & $\times$      & $\times$      & $\times$      & $\times$      \\
    KoDF~\cite{kwon2021kodf}                         & 2021 & 62{,}166/175{,}776       & $\checkmark$  & $\times$      & $\times$      & $\checkmark$  & $\checkmark$  & 6  & $\times$      & $\times$      & $\times$      & $\times$      & $\times$      \\
    FakeAVCeleb-v1.2~\cite{khalid2021fakeavceleb}    & 2021 & 500/19{,}500             & $\checkmark$  & $\times$      & $\times$      & $\checkmark$  & $\checkmark$  & 4  & $\times$      & $\times$      & $\times$      & $\times$      & $\times$      \\
    Celeb-DF-v2~\cite{li2019celeb}                  & 2022 & 590/5{,}639              & $\times$      & $\times$      & $\times$      & $\checkmark$  & $\times$      & 1  & $\times$      & $\times$      & $\times$      & $\times$      & $\times$      \\
    LAV-DF~\cite{cai2022you}                         & 2022 & 36{,}431/99{,}873        & $\checkmark$  & $\times$      & $\times$      & $\times$      & $\times$      & 1  & $\times$      & $\times$      & $\times$      & $\times$      & $\times$      \\
    DF-Platter~\cite{Narayan_2023_CVPR}              & 2023 & 764/132{,}496            & $\checkmark$  & $\times$      & $\times$      & $\checkmark$  & $\checkmark$  & 3  & $\times$      & $\times$      & $\checkmark$  & $\checkmark$  & $\checkmark$  \\
    AV-Deepfake1M~\cite{cai2024av}                   & 2024 & 286{,}721/860{,}039      & $\checkmark$  & $\times$      & $\times$      & $\times$      & $\times$      & 1  & $\times$      & $\times$      & $\times$      & $\times$      & $\times$      \\
    DeepFake-eval-2024~\cite{chandra2025deepfake}    & 2025 & 1{,}072/964              & --            & --            & --            & --            & --            & -- & $\times$      & $\times$      & $\times$      & $\times$      & $\times$      \\
    Celeb-DF++~\cite{li2025celeb}                    & 2025 & 590/53{,}196             & $\checkmark$  & $\checkmark$  & $\checkmark$  & $\checkmark$  & $\checkmark$  & 22 & $\times$      & $\times$      & $\times$      & $\times$      & $\times$      \\
    TalkingHeadBench~\cite{Xiong_2026_WACV}          & 2026 & 2{,}312/2{,}984          & $\checkmark$  & $\checkmark$  & $\times$      & $\times$      & $\checkmark$  & 6  & $\times$      & $\times$      & $\times$      & $\times$      & $\times$      \\
    \midrule
    \textbf{FairTalking-Bench (Ours)}                & \textbf{2026} & \textbf{2{,}964/23{,}712} & $\bm{\checkmark}$ & $\bm{\checkmark}$ & $\bm{\checkmark}$ & $\bm{\checkmark}$ & $\bm{\checkmark}$ & \textbf{8} & $\bm{\checkmark}$ & $\bm{\checkmark}$ & $\bm{\checkmark}$ & $\bm{\checkmark}$ & $\bm{\checkmark}$ \\
    \bottomrule
  \end{tabular}%
  }
  \vspace{-15pt}
\end{table*}
To address these challenges, we introduce \textbf{FairTalking-Bench}, a comprehensive and demographic-balanced video benchmark specifically designed for talking-head deepfake detection. Distinct from previous works, our dataset is constructed based on two guiding principles:~\emph {technological diversity} and \emph{demographic fairness}. We collect high-quality real videos and employ eight cutting-edge generation methods (\emph{e.g.}, Diffusion, AutoEncoder(AE)/GAN, Flow-matching, and 3DMM) to generate 23,712 fake talking-head videos. Crucially, we balance the data across four racial groups, two genders, and six age intervals in 48 balanced demographic subgroups to ensure a demographically balanced evaluation environment. Based on the proposed balanced benchmark, we define four complementary evaluation protocols to assess the robustness, fairness, generalization, and identity bias resilience of detectors. The main contributions of this work are summarized as follows:
\begin{itemize}[leftmargin=*]
    \item To the best of our knowledge, we present the first large-scale, demographically balanced benchmark, namely,~\textbf{FairTalking-Bench}, which consists of demographically balanced talking videos generated by the latest audio-driven talking-head generation methods.

    \item We design four evaluation protocols to systematically assess deepfake detectors under real-world scenarios: zero-shot robustness, demographic fairness, cross-generator generalization, and identity-consistent detection. 

    \item Extensive experiments on our benchmark show orthogonal subspace decomposition and demographic balance improve generalization and fairness. Flow-based training and pretrained representations enhance cross-generator transfer, while generator-specific artifacts and identity-related biases limit generalization across generators.
    
\end{itemize}

\section{Related Work}
\subsection{Deepfake Datasets}
Existing Deepfake video datasets (\emph{e.g.}, UADFV~\cite{UADFV}, FF++~\cite{rossler2019faceforensics++}) are limited by small scales and visible artifacts. While large-scale benchmarks like DFDC~\cite{dolhansky2020deepfake} and ForgeryNet~\cite{he2021forgerynet} significantly improve fidelity and diversity, they tend to neglect demographic balance (age, gender, race). As shown in Table~\ref{tab:deepfake_dataset_demographics}, existing datasets often rely on uncurated web data (\emph{e.g.}, WildDeepfake~\cite{zi2020wilddeepfake}) or focus on specific demographics(\emph{e.g.}, KoDF~\cite{kwon2021kodf})  and manipulation types~\cite{li2020celeb},  primarily face swapping~\cite{DeepfakeTIMIT}. Recent talking-head datasets, such as LAV-DF~\cite{cai2022you}, tend to focus on lip-synchronization and lack fake data generated by diverse generator architectures or fairness guarantees. While DF-Platter~\cite{Narayan_2023_CVPR} provides demographic annotations, it is limited to Indian subjects with only GAN-based face-swapping. Similarly, TalkingHeadBench~\cite{Xiong_2026_WACV} targets talking-head generation yet lacks demographic balance and fairness evaluation protocols.

In contrast to previous works, as highlighted in Table~\ref{tab:deepfake_dataset_demographics}, our benchmark contains large-scale and diverse face forgery videos generated by cutting-edge talking-head generators (including Diffusion~\cite{xu2024hallo}, Flow-matching~\cite{ki2025float}, 3DMM~\cite{zhang2023sadtalker}, and AutoEncoder(AE)/GAN~\cite{tan2024edtalk}), and rigorously ensures a balanced distribution of demographic attributes through the FairFace algorithm and manual verification. It provides ID-paired, demographically balanced, and methodologically diverse talking-head videos, which establish a benchmark for evaluating next-generation detectors.
\subsection{Deepfake Video Generation}
Early research primarily relied on frameworks like FSGAN ~\cite{FSGAN} and StyleGAN~\cite{karras2019style} for face reconstruction. However, these GAN-based methods typically suffer from rigid head movements, distorted expressions, and inconsistent lighting, making them relatively easy tasks for detection.

In recent years, the rapid progress of talking-head generation has catalysed a shift toward high-fidelity synthesis. Emerging diffusion-based models—such as JoyVASA~\cite{cao2024joyvasa}, Hallo~\cite{xu2024hallo}, Sonic~\cite{ji2025sonic}, Ditto~\cite{li2025ditto}, and AniPortrait~\cite{wei2024aniportrait}—integrate audio and reference images to generate temporally coherent videos, effectively mitigating the boundary artifacts prevalent in traditional approaches. This landscape is further enriched by Flow-matching~\cite{ki2025float}, AutoEncoders~\cite{tan2024edtalk}, and 3DMM-based approaches~\cite{zhang2023sadtalker}. The near-authentic quality of generated videos exposes the limitations of existing deepfake datasets, underscoring the urgent need for a novel benchmark specifically tailored for talking-head forgeries to study challenges in identity variations and generator generalization.

\begin{figure*}[t]
  \centering
  \includegraphics[width=\textwidth]{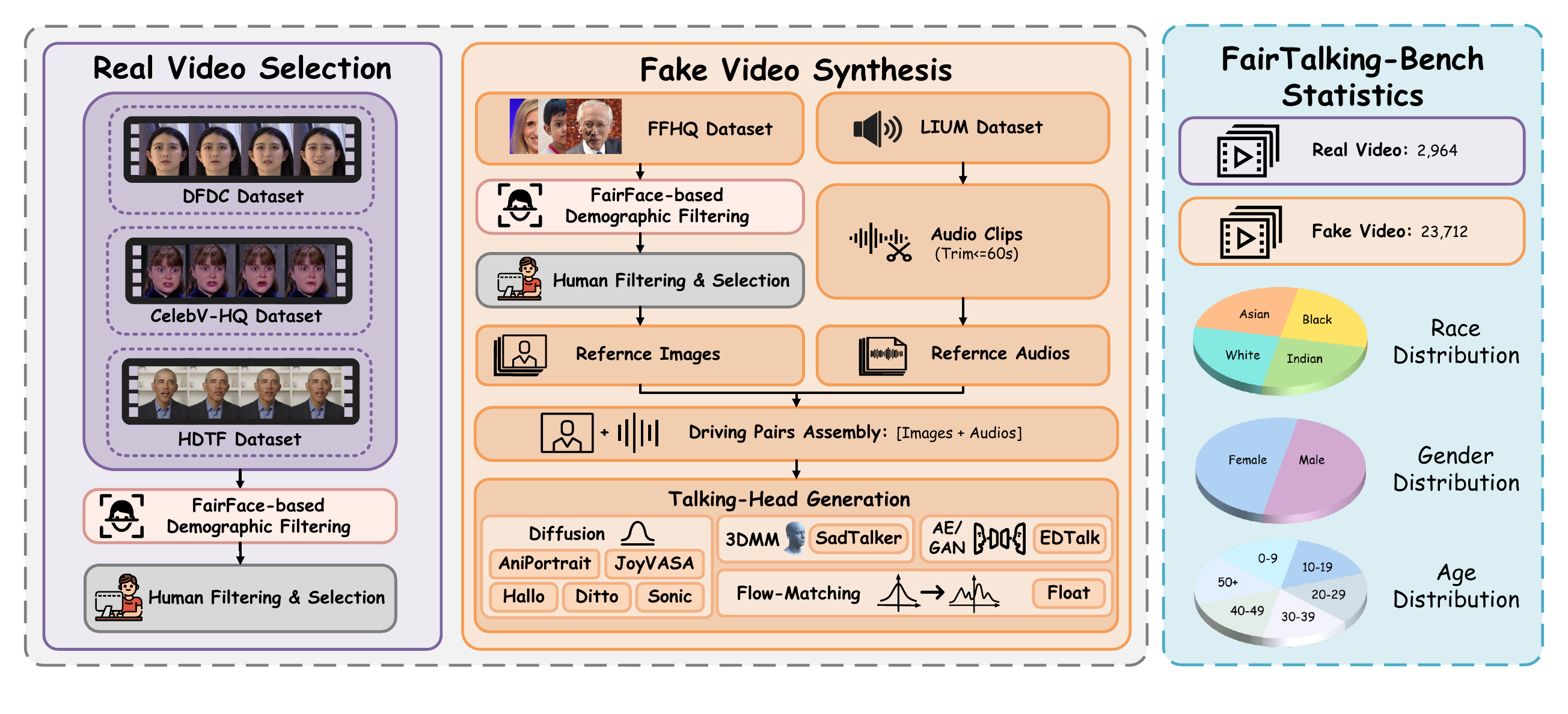}
  \vspace{-25pt}
  \caption{The construction pipeline of FairTalking-Bench.
    The framework consists of: (1) \textbf{Real Video Collection},
    where source videos (DFDC, CelebV-HQ, HDTF) are processed
    through FairFace analysis and human filtering;
    (2) \textbf{Fake Video Synthesis}, which utilizes 2,964
    image-audio driven pairs to generate 23,712 videos via
    eight SOTA models across four paradigms (Diffusion, 3DMM,
    GAN, AE/GAN, and Flow-matching). The final benchmark
    provides a demographic-balanced collection of 26,676
    videos for fair evaluation.}
    
  \label{fig:pipeline}
  \vspace{-10pt}
\end{figure*}

\subsection{Deepfake Video Detection}
Current approaches predominantly leveraged Convolutional Neural Networks (CNNs), such as MesoNet~\cite{afchar2018mesonet} and Xception~\cite{rossler2019faceforensics++}, to focus on low-level texture anomalies and blending boundaries. Some studies enhanced the robustness of detectors through frequency analysis (\emph{e.g.}, RECCE~\cite{cao2022end}), specific lip-motion modeling (LipForensics~\cite{Haliassos_2021_CVPR}), or data augmentation (SBI~\cite{shiohara2022detecting}). To overcome the limited receptive field of CNNs, researchers have introduced Transformer-based architectures. Pure ViT methods like TALL~\cite{xu2023tall} model global dependencies directly from video patches, while hybrid architectures—such as CViT~\cite{wodajo2021deepfake}, FTCN~\cite{Zheng_2021_ICCV}, and LipFD~\cite{liu2024lips}—combine CNN feature extractors with Transformer modules to leverage the complementary nature of local visual details and long-range temporal coherence.

With the advent of large-scale pre-training, recent research has shifted towards foundation models to address the generalization challenges of traditional detectors. Leveraging the rich semantic priors of Vision-Language models (\emph{e.g.}, CLIP), some approaches like Adapter~\cite{cui2025forensics} and DFD-FCG~\cite{Han_2025_CVPR} employ parameter-efficient fine-tuning to adapt general features to forgery detection tasks. Similarly, models like Effort~\cite{yan2024effort} explore orthogonal modelling to mitigate domain shifts. By moving beyond low-level artifacts to incorporate high-level semantic understanding, this paradigm demonstrates potential in detecting unknown forgery types and generalization across diverse datasets.

\section{Dataset Construction}
As illustrated in Figure~\ref{fig:pipeline}, the FairTalking-Bench construction pipeline is divided into two parts: real video collection and fake video synthesis.

\subsection{Real Video Collection}

To curate a collection of real videos, we meticulously filtered high-quality samples from existing public human face video datasets.
Specifically, we exclude datasets characterized by low resolution (\emph{e.g.}, FF++~\cite{rossler2019faceforensics++}), lack of audio tracks (\emph{e.g.}, DeeperForensics-1.0~\cite{jiang2020deeperforensics}), or multi-speaker conversational scenarios (\emph{e.g.}, subsets of DFDC~\cite{dolhansky2020deepfake}), to ensure audio-visual integrity and single-speaker consistency.
Then, to keep demographic taxonomy consistent with related fairness studies(\emph{e.g.}, THB~\cite{Xiong_2026_WACV}, DAG-FDD~\cite{Ju_2024_WACV}), we apply FairFace~\cite{karkkainen2021fairface} to detect demographic characteristics (4 races, 2 genders, 6 age groups) for videos from the remaining datasets: DFDC~\cite{dolhansky2020deepfake}, CelebV-HQ~\cite{zhu2022celebvhq},  and HDTF~\cite{zhang2021flow}.
We further incorporate human verification to ensure reliable demographic labels, 8 annotators are first individually trained until $>$90\% accuracy can be achieved on a held-out validation set. An expert inspector then performs batch-level audits (30\% random sampling per 100-entry batch); batches with error rates exceeding 10\% are fully re-annotated and verified by the expert, ensuring high-quality annotations across all demographic attributes.  Based on these verified labels, we then manually review and select 2,964 high-quality real videos to ensure the balance of their demographic characteristics.

\subsection{Fake Video Synthesis}

For fake video construction, we first assign demographic labels to large -scale, high-quality facial images from FFHQ~\cite{karras2019style} using FairFace~\cite{karkkainen2021fairface}, followed by manual review to ensure label accuracy. We then select 2,964 images that exactly match the balanced demographic distribution of the real collection. Each selected image is paired with a randomly chosen, non-repetitive audio clip ($\leq60$ seconds) from the large-scale, long-talking audio dataset LIUM~\cite{rousseau2014enhancing}. These image-audio pairs are fed into eight representative talking-head generators covering four paradigms: Diffusion(AniPortrait~\cite{wei2024aniportrait}, JoyVASA~\cite{cao2024joyvasa}, Hallo~\cite{xu2024hallo}, Ditto~\cite{li2025ditto}, Sonic~\cite{ji2025sonic}), 3DMM(SadTalker~\cite{zhang2023sadtalker}), Adversarial Autoencoder(EDTalk~\cite{tan2024edtalk}), and Flow-matching(Float~\cite{ki2025float}). This yields 23,712 (2,964 $\times$ 8) synthetic talking-head videos.

\vspace{-2pt}
\subsection{Statistics}
\vspace{-2pt}
    As Figure~\ref{fig:pipeline}~shows, FairTalking-Bench comprises 26,676 videos in total, with 2,964 real and 23,712 synthetic talking-head videos. We enforce demographic balance across both subsets to prevent biases: (1) Race: achieving equal distribution (25\% each) across Asian, White, Black, and Indian groups; (2) Gender: maintaining a balanced 1:1 male-to-female ratio; and (3) Age: ensuring a uniform distribution across six age intervals (0-9 to 50+). The dataset is split into training, validation, and testing sets (8:1:1) with no identity overlap to prevent data leakage. We provide additional benchmark validation in Supp., including demographic annotation verification and video quality screening (Secs.~\textbf{A}), source-distribution analyses (Secs.~\textbf{B}), and cross-language robustness evaluation (Sec.~\textbf{E.2}).

\section{Benchmark Experiments}
\subsection{Evaluation Protocols}
 To systematically assess the robustness, fairness, generalization, and identity bias resilience of SOTA detectors, we design four corresponding experimental protocols:

\vspace{3pt}

\noindent\textbf{Protocol~1: Zero-Shot Robustness.} We evaluate existing pre-trained detectors on FairTalking-Bench without any fine-tuning. This protocol aims to quantify the vulnerability of traditional models when confronting unseen modern generation mechanisms (\emph{e.g.}, Diffusion and Flow-matching) and evaluate baseline performance under domain shifts.

\vspace{3pt}

\noindent\textbf{Protocol~2: Fairness-Aware Evaluation.}
To investigate how demographic distribution affects model fairness, we construct five training subsets ranging from fully balanced to severely imbalanced (\emph{e.g.}, specific racial or gender bias). We train detectors separately using these subsets and evaluate them on the same balanced test set, to investigate whether demographic imbalance leads to spurious correlations and degrades detection performance across demographic groups.

\vspace{3pt}

\noindent\textbf{Protocol~3: Cross-Generator Generalization.}
We evaluate the generalization of detectors through two complementary settings: \textbf{Leave-One-Generator-Out}~\emph{(LOGO)} aims to evaluate the defense capability against unknown attacks by our training data generated by seven generators and testing on the held-out one; \textbf{Leave-One-Generator-In}~\emph{(LOGI)} intends to assess the generalization of detectors by training using one generator and testing on the remaining seven.

\vspace{3pt}

\noindent\textbf{Protocol~4: Identity Consistency Analysis.}
To investigate whether strong detection performance persists when identity differences between real and fake videos are controlled, we evaluate detectors trained on the standard identity-unpaired split using identity-paired videos, in which each fake video shares the identity and driving audio of its source real video. By comparing performance under the standard identity-unpaired and identity-paired settings, this protocol examines sensitivity to identity-related bias and investigates the potential use of identity cues as detection shortcuts.

\subsection{Evaluation Setup}

\noindent\textbf{SOTA Detectors.} We evaluate 17 representative deepfake detectors spanning CNN, Transformer, CNN--Transformer hybrid, and CLIP-based frameworks, as summarized in Table~\ref{tab:EVALUATION}. For zero-shot evaluation, MesoNet, LipFD, and CNNDetection use checkpoints trained on their respective datasets, while CViT and CViT2 use checkpoints trained on DFDC~\cite{dolhansky2020deepfake}. The remaining detectors use official checkpoints pretrained on FaceForensics++~\cite{rossler2019faceforensics++}.

\vspace{3pt}

\noindent\textbf{Evaluation Metrics.} Following Lin et al.~\cite{lin2024preserving}, we employ a comprehensive set of metrics to evaluate both detection performance and demographic fairness.~(1)~\emph{Detection Metrics.}~We report ACC, AUC, TPR@FPR=1\% (T1), and TPR@FPR=0.1\% (T0.1). In Protocol~1, the task is particularly challenging, as existing detectors have never seen these modern talking-head forgeries before and thus struggle to generalize. Consequently, AUC and ACC are sufficient to assess their capabilities. In contrast, for Protocol~2 and 3, where models are trained on our dataset and possess higher discrimination capability, we emphasize stricter thresholds (\emph{e.g.}, T0.1) to evaluate robustness in real-world security scenarios. (2)~\emph{Fairness Metrics.} To quantify the disparity in detection performance across different demographic groups, we adopt four fairness metrics: Equal False Positive Rate ($F_{\text{FPR}}$), Max Equalized Odds ($F_\text{{MEO}}$), Demographic Parity ($F_{\text{DP}}$), and Overall Accuracy Equality ($F_{\text{OAE}}$). Lower values indicate better fairness. Detailed metric definitions, demographic subgroup analyses, and bootstrap confidence intervals are provided in \textit{Supp.~\textbf{C}} 

\begin{table*}[t]
    \centering
    \caption{Protocol 1: zero-shot robustness on the FairTalking-Bench
    testing set. Best and second-best results are in \textbf{bold}
    and \underline{underlined}. \textbf{Trans.}: Transformer;
    \textbf{Mod.}: modality; \textbf{V}: video; \textbf{I}: image.}
    \label{tab:EVALUATION}
    \scriptsize
    \renewcommand{\arraystretch}{1.05}
    \setlength{\tabcolsep}{3pt}
    \begin{tabular*}{\textwidth}{@{\extracolsep{\fill}}llllcrrrr@{}}
        \toprule
        \textbf{Detector} & \textbf{Framework} & \textbf{Venue}
        & \textbf{Training dataset} & \textbf{Mod.}
        & \textbf{AUC} $\uparrow$ & \textbf{ACC} $\uparrow$
        & \textbf{FLOPs (G)} & \textbf{Param. (M)} \\
        \midrule
        Meso~\cite{afchar2018mesonet}
        & \multirow{8}{*}{CNN} & WIFS18 & Custom dataset & V
        & 60.57 & 69.13 & 0.06 & 0.03 \\
        Xception~\cite{rossler2019faceforensics++}
        & & ICCV19 & FaceForensics++ & V
        & 59.74 & 45.31 & 6.01 & 20.81 \\
        CNNDetection~\cite{wang2020cnn}
        & & CVPR20 & Custom dataset & I
        & 45.94 & 11.15 & 4.13 & 23.51 \\
        Lipforensic~\cite{Haliassos_2021_CVPR}
        & & CVPR21 & FaceForensics++ & V
        & 59.56 & 23.61 & 0.72 & 35.99 \\
        RECCE~\cite{cao2022end}
        & & CVPR22 & FaceForensics++ & I
        & 62.38 & 43.81 & 22.58 & 25.83 \\
        SBI~\cite{shiohara2022detecting}
        & & CVPR22 & FaceForensics++ & I
        & \textbf{81.04} & 62.47 & 9.02 & 17.55 \\
        AltFreezing~\cite{Wang_2023_CVPR}
        & & CVPR23 & FaceForensics++ & V
        & \underline{76.76} & 55.69 & 7.13 & 27.23 \\
        ProDet~\cite{cheng2024can}
        & & NeurIPS24 & FaceForensics++ & V
        & 70.58 & \underline{83.30} & 3.63 & 55.98 \\
        \hdashline
        UIA-ViT~\cite{zhuang2022uia}
        & \multirow{2}{*}{Trans.} & ECCV22 & FaceForensics++ & I
        & 44.25 & 21.87 & 35.17 & 85.80 \\
        TALL~\cite{xu2023tall}
        & & ICCV23 & FaceForensics++ & V
        & 35.75 & \textbf{88.54} & 4.09 & 86.92 \\
        \hdashline
        CViT~\cite{wodajo2021deepfake}
        & \multirow{4}{*}{CNN+Trans.} & arXiv21 & DFDC & V
        & 38.87 & 18.19 & 13.33 & 90.04 \\
        FTCN~\cite{Zheng_2021_ICCV}
        & & ICCV21 & FaceForensics++ & V
        & 64.72 & 33.63 & 4.21 & 14.77 \\
        LipFD~\cite{liu2024lips}
        & & NeurIPS24 & Custom dataset & V
        & 46.51 & 15.20 & 57.09 & 327.51 \\
        CViT2~\cite{wodajo2024deepfake}
        & & GEM24 & DFDC & V
        & 44.14 & 22.25 & 13.33 & 90.04 \\
        \hdashline
        Adapter~\cite{cui2025forensics}
        & \multirow{3}{*}{CLIP} & CVPR25 & FaceForensics++ & I/V
        & 59.74 & 30.14 & 167.58 & 435.16 \\
        DFD-FCG~\cite{Han_2025_CVPR}
        & & CVPR25 & FaceForensics++ & V
        & 70.20 & 41.36 & 162.79 & 303.43 \\
        Effort~\cite{yan2024effort}
        & & ICML25 & FaceForensics++ & I
        & 71.24 & 62.85 & 51.95 & 202.16 \\
        \bottomrule
    \end{tabular*}
    \vspace{-5pt}
\end{table*}

\vspace{3pt}

\subsection{Protocol 1: Zero-shot Robustness}

\noindent\textbf{Generalization-oriented designs show promise for zero-shot transfer.} Table~\ref{tab:EVALUATION} highlights the challenges of zero-shot detection on FairTalking-Bench, with 15 of 17 detectors achieving AUCs below 72\%. SBI~\cite{shiohara2022detecting} and AltFreezing~\cite{Wang_2023_CVPR} achieve the highest AUCs of 81.04\% and 76.76\%, respectively. Their stronger generalization may be attributed to training strategies that reduce reliance on manipulation-specific artifacts. Specifically, SBI introduces diverse pseudo-forgeries through self-blending, whereas AltFreezing alternates the optimization of spatial and temporal representations.

{\noindent\textbf{Orthogonal subspace decomposition may improve zero-shot generalization of CLIP-based detectors.} Although Adapter~\cite{cui2025forensics}, DFD-FCG ~\cite{Han_2025_CVPR}, and Effort ~\cite{yan2024effort} share the same CLIP backbone, Effort with orthogonal subspace decomposition achieves the highest AUC (71.24\%), outperforming CLIP pre-training-based Adapter (59.74\%) and parameter-efficient fine-tuning-based DFD-FCG (70.20\%). This suggests that CLIP pre-training provides transferable semantic representations, while parameter-efficient fine-tuning improves adaptation to manipulation-specific features. However, orthogonal subspace decomposition better balances the preservation of transferable semantic knowledge and the learning of forgery-related features, leading to improved zero-shot generalization.}

\vspace{3pt}

\noindent\textbf{Lip-focused detectors may exhibit limited generalization to modern talking-head synthesis.}
LipForensics~\cite{Haliassos_2021_CVPR} and LipFD~\cite{liu2024lips} achieve AUCs of only 59.56\% and 46.51\%, respectively. This performance degradation may arise from differences in lip-motion artifacts between their training data and modern talking-head synthesis. As modern generators produce increasingly coherent lip movements, conventional lip-based temporal cues become less discriminative, highlighting the importance of integrating complementary spatial and temporal representations.

\vspace{3pt}

\noindent\textbf{Higher computational cost does not imply better zero-shot performance.}
SBI achieves the highest AUC with only 17.55M parameters and 9.02G FLOPs, whereas Adapter requires 435.16M parameters and 167.58G FLOPs but obtains an AUC of 59.74\%. Similarly, LipFD contains 327.51M parameters yet obtains an AUC of only 46.51\%. These results suggest that increased model complexity alone does not guarantee stronger generalization to unseen talking-head forgeries.

Together, these results show that FairTalking-Bench provides a challenging benchmark for evaluating the cross-domain generalization of both conventional and foundation-model-based approaches.

\subsection{Protocol 2: Fairness-Aware Evaluation}
We compare Xception, FTCN, TALL, and Adapter under balanced and imbalanced training settings. Each training set includes 2,371 real and 2,372 fake videos, whose demographic composition differs between the two settings. We keep the real and fake identity pools distinct and evaluate models on the same balanced test set, which allows us to compare demographic sampling settings at a fixed training-set size and real/fake ratio.Additional comparisons across five training subsets with varying demographic distributions and sample counts are provided in \textit{Supp.} (Secs.~\textbf{C}). 


\begin{table}[t]
    \centering
    \caption{Detection and fairness
performance (\%) on balanced demographic testing
sets. \textbf{Bold} indicates the better result.}
\vspace{-3pt}
    \label{tab:matched_4743}
    \scriptsize
    \setlength{\tabcolsep}{3pt}
    \renewcommand{\arraystretch}{1.05}
    \resizebox{\linewidth}{!}{%
    \begin{tabular}{llrrrrrrrr}
        \toprule
        \multirow{2}{*}{Detector}
        & \multirow{2}{*}{Training set}
        & \multicolumn{4}{c}{Detection metrics (\%) $\uparrow$}
        & \multicolumn{4}{c}{Fairness metrics (\%)$\downarrow$} \\
        \cmidrule(lr){3-6}\cmidrule(lr){7-10}
        & & AUC & T1 & T0.1 & ACC
        & $F_{\mathrm{FPR}}$ & $F_{\mathrm{OAE}}$
        & $F_{\mathrm{DP}}$ & $F_{\mathrm{MEO}}$ \\
        \midrule
        \multirow{2}{*}{Xception}
        & Balanced
        & \textbf{99.98} & \textbf{99.53} & \textbf{98.53} & \textbf{99.35}
        & \textbf{41.05} & 2.47 & \textbf{5.63} & \textbf{22.22} \\
        & Imbalanced
        & 99.50 & 97.52 & 87.89 & 99.06
        & 51.11 & \textbf{2.46} & 7.14 & 29.17 \\
        \midrule
        \multirow{2}{*}{FTCN}
        & Balanced
        & \textbf{99.54} & \textbf{90.00} & \textbf{37.03} & \textbf{98.20}
        & \textbf{63.55} & 3.39 & \textbf{5.94} & \textbf{30.95} \\
        & Imbalanced
        & 99.00 & 75.45 & 29.44 & 97.20
        & 77.19 & \textbf{3.05} & 6.35 & 49.46 \\
        \midrule
        \multirow{2}{*}{TALL}
        & Balanced
        & \textbf{99.92} & \textbf{97.11} & \textbf{89.99} & \textbf{99.50}
        & \textbf{27.33} & \textbf{1.24} & \textbf{5.63} & \textbf{11.11} \\
        & Imbalanced
        & 98.97 & 88.59 & 67.95 & 97.43
        & 49.38 & 3.53 & 5.75 & 22.22 \\
        \midrule
        \multirow{2}{*}{Adapter}
        & Balanced
        & \textbf{99.99} & \textbf{99.91} & \textbf{97.75} & \textbf{99.77}
        & \textbf{14.34} & \textbf{0.62} & \textbf{5.63} & \textbf{5.56} \\
        & Imbalanced
        & 99.72 & 95.52 & 79.88 & 96.54
        & 29.80 & 1.78 & 5.64 & 5.67 \\
        \bottomrule
    \end{tabular}%
    }
    \vspace{-15pt}
\end{table}

\noindent\textbf{Demographic balance improves detection performance and fairness.} In the table~\ref{tab:matched_4743}, detectors trained on demographically balanced subsets generally achieve stronger detection performance and reduced fairness disparities than those trained on imbalanced subsets. For example, balanced population distribution improves TALL's AUC from 98.97\% to 99.92\% and T0.1 from 67.95\% to 89.99\%, while reducing its $F_{\mathrm{FPR}}$ from 49.38\% to 27.33\% and $F_{\mathrm{MEO}}$ from 22.22\% to 11.11\%. Similar trends are observed across other detectors, suggesting that demographic balance in training data can benefit both detection performance and fairness.

\vspace{3pt}
\noindent\textbf{Better fairness is generally associated with stronger detection performance.} In the table~\ref{tab:matched_4743}, detectors trained on demographically balanced subsets generally achieve both lower fairness disparities and stronger detection performance than those trained on imbalanced subsets. For example, among all detectors, Adapter achieves the lowest $F_{\text{MEO}}$ of 5.56\% and the highest T0.1 of 97.75\% under balanced training, suggesting a positive association between fairness and detection performance.

\vspace{3pt}

\noindent\textbf{Pretrained representation adaptation may facilitate accurate and fair detection.}
Under balanced training, Adapter achieves the lowest or joint-lowest values across all four disparity metrics, together with the highest AUC and ACC and a T0.1 of 97.75\%. This suggests that adapting pretrained CLIP representations can preserve transferable visual knowledge while learning forgery-specific cues, offering a promising balance between detection performance and demographic fairness.

\vspace{3pt}

\noindent\textbf{Low-FPR evaluation reveals performance differences masked by saturated AUC.} Adapter achieves near-saturated AUC under both training conditions, yet its low-FPR sensitivity differs substantially. Balanced training increases AUC from 99.72\% to 99.99\%, a gain of only 0.27\% , while improving T0.1 from 79.88\% to 97.75\%, a gain of 17.87\%. These results highlight the importance of low-FPR evaluation in revealing the benefits of demographic balance beyond aggregate AUC. We analyze interactions between demographic attributes and generators at different operating points using shared decision thresholds in Supp. (Sec.~\textbf{C.8}).


\begin{figure*}
    \centering
    \includegraphics[width=\linewidth]{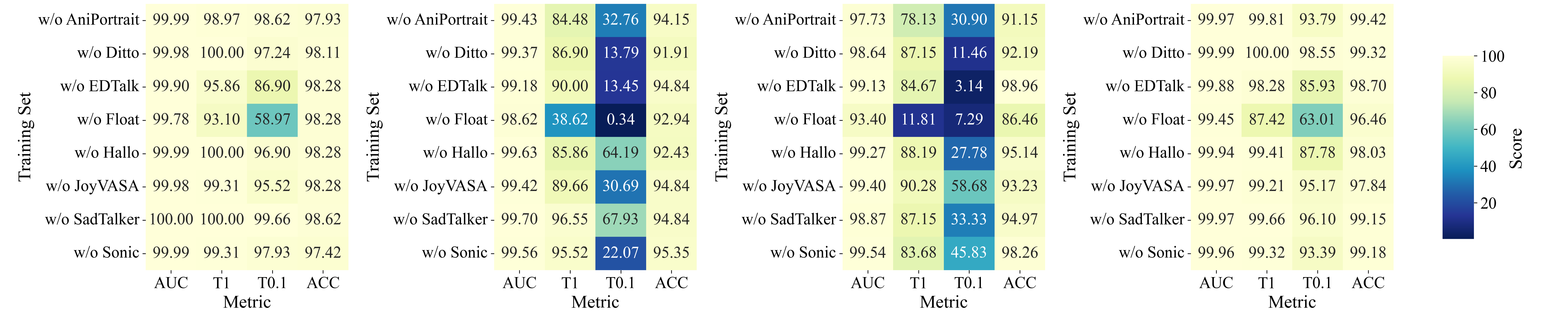}
    \caption{Protocol 3-LOGO performance heatmaps for Xception, FTCN, TALL, and Adapter, from left to right.}
    \label{fig:train7_LOGO_combined}
     \vspace{-15pt}
\end{figure*}

\begin{figure*}
    \centering
    \includegraphics[width=\linewidth]{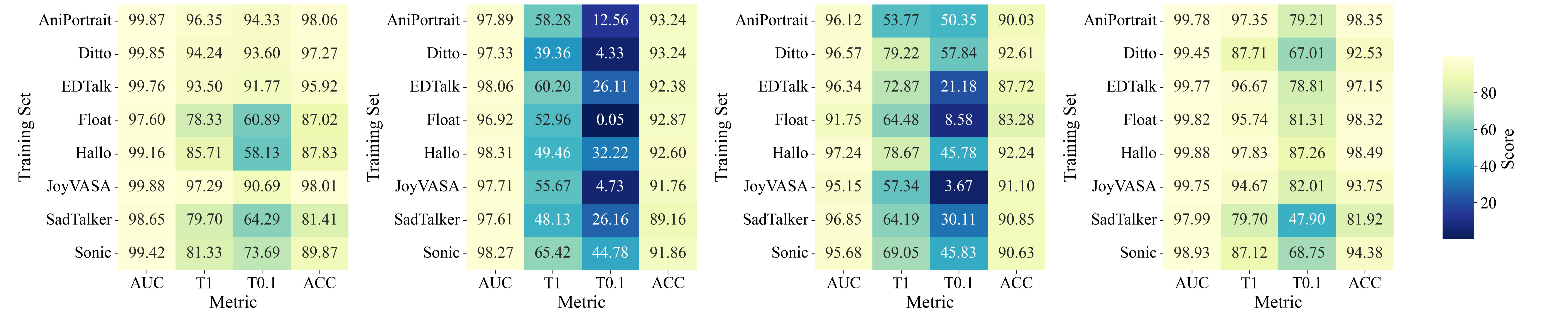}
    \caption{Protocol 3-LOGI performance heatmaps for Xception, FTCN, TALL, and Adapter, from left to right.}
    \label{fig:train1_LOGI_combined}
     \vspace{-15pt}
\end{figure*}

\vspace{3pt}

\subsection{Protocol 3: Cross-Generator Generalization}
To evaluate the generalization performance against unseen data, we employ two complementary settings: \textit{Leave-One-Generator-Out (LOGO)} and \textit{Leave-One-Generator-In (LOGI)}. These settings challenge models to distinguish universal forgery traces from specific generator artifacts.
Building on Protocol 2's finding that demographic balance is essential for robust low-FPR detection, as imbalanced training data introduces spurious correlations that critically limit real-world deployment.

\vspace{3pt}

\noindent\textbf{Flow-based generators may pose significant challenges to detectors.} As illustrated in Figure~\ref{fig:train7_LOGO_combined}, most detectors achieve poorer performance on flow-based generators than on diffusion-based ones. This suggests that flow-based generators may produce more realistic synthetic faces with fewer detectable artifacts, making them more challenging for existing detectors to identify forgery traces.

\noindent\textbf{Pretrained representations may facilitate cross-generator generalization.}
Adapter achieves a T0.1 exceeding 85\% across all LOGO settings except when Float is held out, and attains 81.31\% when trained solely on Float in LOGI. These results suggest that pretrained CLIP representations, combined with lightweight adaptation, facilitate the learning of transferable forgery cues while mitigating reliance on generator-specific artifacts.

\noindent\textbf{Generator-specific artifacts may limit cross-generator generalization.} The LOGI analysis  in Figure~\ref{fig:train1_LOGI_combined} demonstrates that detectors are trained using a generator, its performance degrades substantially when evaluated on seven unseen generators, where the demographic distributions across all test generators remain consistent. This indicates that the cross-generator performance degradation is mainly associated with forgery-agnostic generator-specific artifacts rather than demographic features.

\noindent\textbf{Flow-based training distributions tend to enhance cross-generator generalization.} As shown in Figure ~\ref{fig:train7_LOGO_combined} and~\ref{fig:train1_LOGI_combined}, detectors trained using flow-based generators exhibit stronger cross-generator generalization than those trained with diffusion-based methods. This suggests that flow-based training data may facilitate the learning of more transferable forgery representations, thereby reducing reliance on generator-specific artifacts and improving robustness across diverse generation paradigms.


\subsection{Protocol 4: Identity Consistency Analysis}

\begin{table}[t]
    \centering
    \caption{Protocol 4: identity consistency analysis on
    FairTalking-Bench. Best and second-best results within each testing subset are in \textbf{bold} and \underline{underlined}.}
    \vspace{-5pt}\label{tab:fairtalkingbench_performance}
    \scriptsize
    \setlength{\tabcolsep}{2pt}
    \renewcommand{\arraystretch}{1.05}
    \begin{tabular*}{\linewidth}{@{\extracolsep{\fill}}lllrrrr@{}}
        \toprule
        \multirow{2}{*}{\shortstack{\textbf{Training}\\\textbf{set}}}
        & \multirow{2}{*}{\shortstack{\textbf{Testing}\\\textbf{set}}}
        & \multirow{2}{*}{\textbf{Detector}}
        & \multicolumn{4}{c}{\textbf{Metrics}} \\
        \cmidrule(lr){4-7}
        & & & AUC$\uparrow$ & T1$\uparrow$
        & T0.1$\uparrow$ & ACC$\uparrow$ \\
        \midrule
        \multirow{12}{*}{\shortstack{FairTalking-\\Bench\\(Balanced)}}
        & \multirow{4}{*}{Subset A}
        & Xception & \underline{51.95} & 0.60
        & 0.60 & \textbf{80.32} \\
        & & FTCN & 38.95 & 1.80
        & \underline{1.80} & 43.09 \\
        & & TALL & 44.91 & \underline{3.57}
        & \textbf{3.57} & 15.96 \\
        & & Adapter & \textbf{57.87} & \textbf{3.73}
        & 0.43 & \underline{59.34} \\
        \cmidrule(lr){2-7}
        & \multirow{4}{*}{Subset B}
        & Xception & 46.24 & \textbf{1.89}
        & \textbf{1.89} & \textbf{57.44} \\
        & & FTCN & 42.20 & 0.47
        & \underline{0.47} & \underline{42.98} \\
        & & TALL & \underline{46.44} & 0.00
        & 0.00 & 27.31 \\
        & & Adapter & \textbf{54.54} & \underline{1.16}
        & 0.26 & 24.69 \\
        \cmidrule(lr){2-7}
        & \multirow{4}{*}{Subset C}
        & Xception & \underline{47.71} & 0.48
        & \underline{0.06} & \textbf{55.47} \\
        & & FTCN & 44.59 & \textbf{0.93}
        & 0.03 & \underline{40.76} \\
        & & TALL & 45.96 & 0.67
        & 0.00 & 22.91 \\
        & & Adapter & \textbf{53.82} & \underline{0.89}
        & \textbf{0.12} & 24.71 \\
        \bottomrule
    \end{tabular*}
    \vspace{-15pt}
\end{table}
We evaluate detectors on three testing subsets: \textit{Subset A} (Training IDs), \textit{Subset B} (Testing IDs), and \textit{Subset C} (Unseen IDs), with all IDs drawn from the HDTF dataset in our benchmark.

\vspace{3pt}

\noindent\textbf{Identity-paired settings significantly challenge detection generalization.}
As shown in Table~\ref{tab:fairtalkingbench_performance}, existing detectors exhibit substantial performance degradation under identity-paired settings, with AUCs ranging from 38.95\% to 57.87\%, 42.20\% to 54.54\%, and 44.59\% to 53.82\% on Subsets A, B, and C, respectively. Notably, Xception's AUC drops from 99.98\% under the standard balanced setting to 46.24\% on Subset B, a decrease of 53.74\%. Moreover, performance remains consistently low across all three subsets, even on Subset A containing training identities. These findings suggest that identity familiarity alone is insufficient to ensure reliable detection under identity-paired synthesis, highlighting the limited generalization of existing detectors to this challenging setting.

\vspace{3pt}

\begin{figure}[t]
    \centering
    \includegraphics[width=\linewidth]{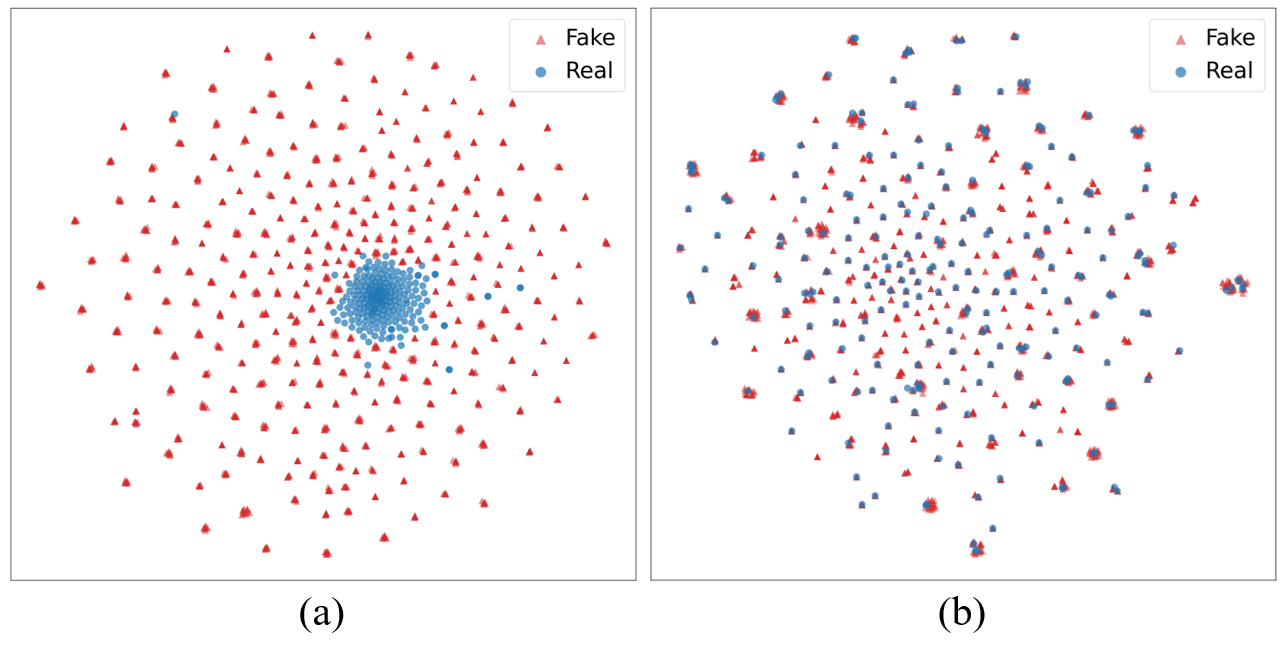}
    \caption{Visualization of identity features extracted by ArcFace:
    \textbf{(a)} the standard FairTalking-Bench testing set;
    \textbf{(b)} the identity-paired dataset for Protocol 4.}
    \label{fig:id-paird}
    \vspace{-15pt}
\end{figure}

\vspace{3pt}

\noindent\textbf{Identity-related shortcuts are a plausible explanation.}
To validate this hypothesis, we visualize the identity feature distributions extracted from our identity-unpaired testing set and the identity-paired one using ArcFace~\cite{deng2019arcface}. As shown in Fig.~\ref{fig:id-paird}(a), there are obvious identity distribution distinctions between real and fake faces in the identity-unpaired testing set, i.e., identity features between real and fake samples are clustered in different domains. This phenomenon may facilitate the detectors to identify fake videos through identity information rather than forgery artifacts. In contrast, as Fig.~\ref{fig:id-paird}(b) shows, with the introduction of identity-paired data, the identity distributions between real and fake faces cluster together and form a similar distribution. As Table~\ref{tab:fairtalkingbench_performance} displays, while the identity bias in the testing set is eliminated, the ability of detectors to distinguish deepfakes is effectively degraded. This demonstrates that current detectors may learn to use identity information for deepfake detection, highlighting the need to disentangle identity from forgery clues. Although modern generators may introduce heterogeneous synthesis patterns. However, the goal of Protocol 4 is not to attribute failures to generator-specific low-level artifacts, but to investigate whether detectors learn forgery-specific representations or identity-related shortcuts. By controlling identity consistency between real and fake videos, Protocol 4 removes identity discrepancy while keeping other generative settings, including the generator, unchanged. Therefore, the observed degradation suggests that existing detectors partially rely on identity-related cues rather than general forgery-specific artifacts.We provide a controlled paired/unpaired comparison in Supp. (Sec.~\textbf{D.1}), where both settings use the same real HDTF videos and driving audio and differ only in the reference images used for synthesis.

\vspace{3pt}

\noindent\textbf{The performance reduction extends to independently captured id-paird videos.}
To further verify the generalizability of this observation, we collect an independent identity-paired evaluation set from 126 participants. Each participant provides three talking-head videos under different scenarios, while keeping the same control variables as the original Protocol~4 setting. In total, we construct a larger identity-paired evaluation set containing 3,402 videos, including 378 real videos and 3,024 fake videos generated using the same talking-head generation pipeline. We evaluate the detectors under this independently captured setting. The results remain consistent with the original observation: all four detectors exhibit low discrimination, achieving AUCs of 45.36\% (Xception), 37.78\% (FTCN), 45.92\% (TALL), and 50.19\% (Adapter). These results show that the performance degradation persists in the larger participant collection, supporting the observation that existing detectors struggle when real and fake videos share the same identity. Thus, detectors may rely on identity-related cues as shortcuts rather than intrinsic forgery artifacts, highlighting the need to disentangle identity from forgery clues.

\subsection{User Study}
To examine the quality of our dataset, we conduct two complementary studies. \textbf{(1) Muted visual study:} Since Celeb-DF++~\cite{li2025celeb} and FF++~\cite{rossler2019faceforensics++} lack audio tracks, we first evaluate visual fidelity alone. 30 participants rate 24 randomly sampled video sets from four datasets on a 1--4 scale (muted). As shown in Table~\ref{tab:userStudy}, our dataset achieves the highest average score(2.83). While FF++ also scores relatively high, this is attributed to its face-swapping approach preserving authentic backgrounds and occupying smaller frame areas, effectively masking artifacts. In contrast, the close-up nature of talking-head videos exposes artifacts more directly, yet our dataset still outperforms all competitors. \textbf{(2) Audiovisual study:} Since talking-head deepfakes are inherently audiovisual, we further conduct a full audio-visual evaluation. 30 participants compare 12 paired video sets between our dataset and THB~\cite{Xiong_2026_WACV}: 76.77\% preferred our dataset for visual realism (natural expressions and head movements), and 69.70\% for lip-audio synchronization. These results robustly validate the superior audiovisual fidelity of FairTalking-Bench.

\begin{table}[t]
    \centering
    \caption{User study results. \textbf{(Left)} Muted visual study,\textbf{(Right)} Audiovisual study,S: score; A: average.}
    \label{tab:userStudy}
    \scriptsize
    \setlength{\tabcolsep}{3pt}
    \renewcommand{\arraystretch}{1.05}
    \resizebox{\linewidth}{!}{%
    \begin{tabular}{lccccccc}
        \toprule
        & \multicolumn{5}{c}{\textbf{Muted visual study}}
        & \multicolumn{2}{c}{\textbf{Audiovisual study}} \\
        \cmidrule(lr){2-6}\cmidrule(lr){7-8}
        \textbf{Dataset}
        & \textbf{S1} & \textbf{S2} & \textbf{S3} & \textbf{S4}
        & \textbf{A}
        & \shortstack{\textbf{Visual}\\\textbf{Realism}}
        & \shortstack{\textbf{Lip-Audio}\\\textbf{Sync.}} \\
        \midrule
        FF++~\cite{rossler2019faceforensics++}
        & 31.53\% & 10.42\% & 18.33\% & 39.72\%
        & \underline{2.66} & -- & -- \\
        CelebDF++~\cite{li2025celeb}
        & 16.25\% & 33.33\% & 34.03\% & 16.39\%
        & 2.51 & -- & -- \\
        THB~\cite{Xiong_2026_WACV}
        & 39.58\% & 27.78\% & 25.00\% & 7.64\%
        & 2.01 & 23.23\% & 30.30\% \\
        \midrule
        Ours
        & 12.64\% & 28.47\% & 22.64\% & 36.25\%
        & \textbf{2.83} & \textbf{76.77\%} & \textbf{69.70\%} \\
        \bottomrule
    \end{tabular}%
    }
    \vspace{-10pt}
\end{table}

\subsection{Visualization}
\begin{figure}[t]
    \centering
    \includegraphics[width=\columnwidth]{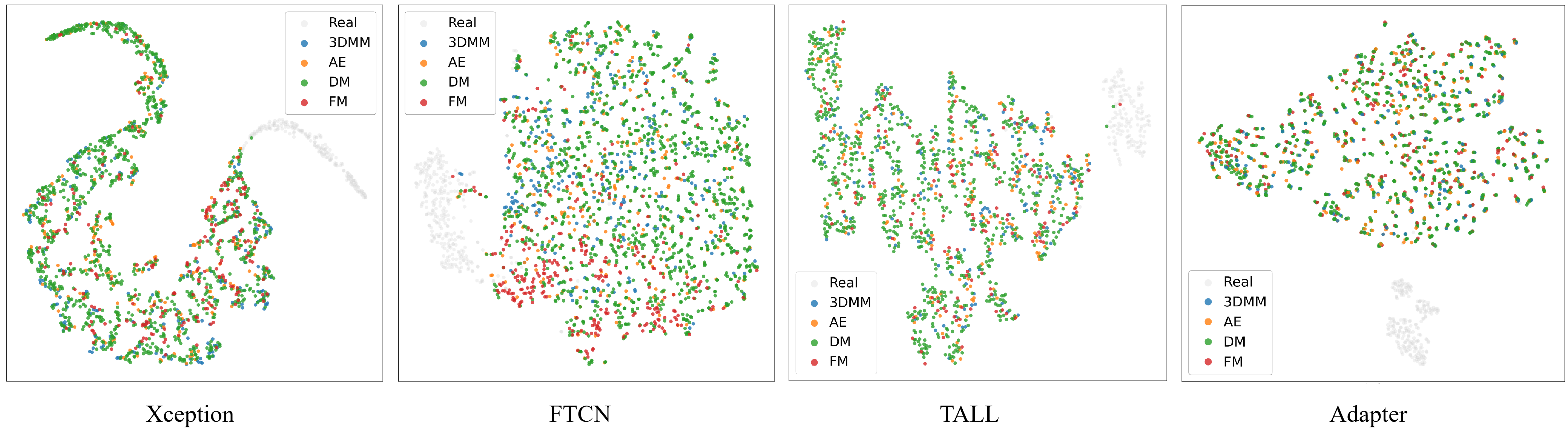}
    \caption{t-SNE visualization of feature spaces extracted by Xception, FTCN, TALL, and FA. Different colors represent real samples and forgery paradigms, respectively.}
    \label{fig:t-SNE_framework}
    \vspace{-10pt}
\end{figure}

As illustrated in Figure~\ref{fig:t-SNE_framework}, after training on the balanced dataset, the t-SNE visualization of features extracted from the testing set by various methods demonstrates a relatively clear global separability between real and fake samples. We notice that the authentic and forged features captured by various methods are distributed into relatively independent regions within the feature space, indicating that these methods possess a certain degree of discriminative capability for binary forgery detection.

However, as shown in Figure~\ref{fig:t-SNE_framework}, when examining the discrimination among different forgery categories, a distinct limitation emerges: although models maintain the separation between real and fake classes, the feature distributions exhibit a high degree of overlap across different forgery methods. Specifically, forged samples generated by different methods fail to form cluster structures; instead, they present a mixed and entangled distribution. This shows that current models struggle to effectively capture fine-grained differences between diverse forgery techniques.

\section{Conclusion}

    We introduce FairTalking-Bench, a demographically balanced talking-head deepfake detection benchmark with diverse synthesis methods and four complementary evaluation protocols. Extensive experiments show that orthogonal subspace decomposition tends to facilitate zero-shot detection. Balanced demographic sampling improves detection performance and fairness. Flow-based generators challenge detectors but offer promising training distributions for improved cross-generator generalization. Pretrained representations promote cross-generator generalization, whereas generator-specific artifacts hinder it. Identity-paired evaluation reveals identity-related shortcuts. Overall, shortcut behaviours are influenced not only by generation paradigms but also by identity information, suggesting that detectors may exploit identity-related cues or generator-specific artifacts instead of purely forgery-specific evidence. This further highlights the importance of disentangling identity information, generator-specific artifacts, and forgery-related features for robust and fair detection.

\section*{LLM Usage Statement}
No large language models (LLMs) were used in the preparation of this manuscript.

{
    \small
    \bibliographystyle{ieeenat_fullname}
    \bibliography{main}
}

\clearpage
\setcounter{page}{1}
\maketitlesupplementary
\appendix

\setcounter{figure}{0}
\setcounter{table}{0}
\renewcommand{\thefigure}{S\arabic{figure}}
\renewcommand{\thetable}{S\arabic{table}}

\section{Dataset Construction Details}
\label{sec:dataset_details} 
\subsection{Why Rebalancing Existing Datasets Is Insufficient}
Simply rebalancing existing datasets is insufficient for three reasons:

\begin{itemize}
    \item \textbf{Lack of comprehensive demographic annotations:} DFDC, CelebV-HQ, HDTF, and Celeb-DF++ lack sample-level (race, gender, age) triplets; DF-Platter provides annotations but covers only a single racial group (Indian).

    \item \textbf{Insufficient coverage of modern generation paradigms:}Existing datasets mainly focus on traditional face manipulation methods, such as face swapping, while novel manipulations like talking-head generation are rarely included. Moreover, they predominantly contain GAN-based forgeries, with limited coverage of diffusion-based and flow-matching-based methods. Therefore, rebalancing existing datasets can improve demographic balance within existing generator categories, but cannot simultaneously ensure comprehensive demographic diversity and coverage of advanced forgery paradigms.

    \item \textbf{Infeasibility of identity-paired evaluation:}Existing datasets generally do not provide an identity-paired structure in which real and fake samples share the same identity and audio sources. Without the required identity and audio correspondence, this protocol cannot be evaluated through simple sampling or attribute-based rebalancing.
\end{itemize}

By contrast, FairTalking-Bench provides talking-head videos generated by \textbf{eight advanced generators} spanning diffusion, flow matching, AE/GAN, and 3DMM, with demographic balance across \textbf{48 subgroups}.

\subsection{Real Video Collection}
Our real videos are sourced from three publicly available datasets: CelebV-HQ~\cite{zhu2022celebvhq}, DFDC~\cite{dolhansky2020deepfake}, and HDTF~\cite{zhang2021flow}. To ensure data quality and consistency, we apply the following exclusion criteria: (i)~videos without accompanying audio tracks are removed;
\begin{enumerate}[leftmargin=*]
      \item \textbf{CelebV-HQ~\cite{zhu2022celebvhq}:} This dataset is a large-scale, high-quality collection of video facial attributes. It comprises 35,666 video clips involving 15,653 identities, with manual annotations for 83 facial attributes, including 40 appearance attributes, 35 action attributes, and 8 emotion attributes. The videos are at a resolution of 512 × 512, with a total duration of approximately 63 hours. The dataset can be accessed through the following link: \url{https://github.com/CelebV-HQ/CelebV-HQ}. The license information for this dataset is not explicitly mentioned.
      \item \textbf{DFDC~\cite{dolhansky2020deepfake}:} This large-scale dataset was utilized in the DeepFake Detection Challenge (DFDC) Kaggle competition. It contains over 100,000 video clips from 3,426 paid actor, produced with several Deepfake, GAN-based, and non-learned methods. The dataset can be accessed through the following link: \url{https://www.kaggle.com/c/deepfake-detection-challenge}. The license information can be found at \url{https://www.kaggle.com/competitions/deepfake-detection-challenge/rules#7-competition-data}.
      \item \textbf{HDTF~\cite{zhang2021flow}:} This dataset was collected from YouTube, comprising approximately 16 hours of 720p or 1080p talking-head videos. A keypoint detector was employed to crop the face regions from the source videos. Each cropped video was then resized to a resolution of 512 × 512, while maintaining high visual quality. The dataset can be accessed through the following link: \url{https://github.com/MRzzm/HDTF} under Creative Commons Attribution 4.0 International License.
  \end{enumerate}
\subsection{Source Datasets and Assets}
\paragraph{Source Images for Fake Video Generation.}Our dataset uses facial images from FFHQ (Flickr Faces High Quality)~\cite{karras2019style}. It is a high-definition facial dataset that contains 70000 facial images with a resolution of 1024 × 1024. Different people have significant differences in features such as race and age, and cover a wide range. Facial images with different attribute features can be selected from it. The facial images in FFHQ are more realistic, presenting rich and diverse backgrounds. Different images are very flexible and versatile in terms of facial expressions, head postures, hairstyles, and accessory decorations (such as glasses, hats, headscarves, etc.). The dataset can be accessed through the following link: \url{https://github.com/NVlabs/ffhq-dataset} under Creative Commons BY-NC-SA 4.0 license.

\paragraph{Source Driving Audio for Fake Video Generation.}The driving signals used in our dataset are all audio-based, sourced from the TED-LIUM corpus (LIUM)~\cite{rousseau2014enhancing}. This dataset is an English speech recognition dataset based on TED talks, designed specifically for Automatic Speech Recognition (ASR) tasks. The data sampling rate is 16kHz, including audio files and their transcribed text. We only use the audio file, which are all in .wav format. The dataset can be accessed through the following link: \url{https://huggingface.co/datasets/LIUM/tedlium} under CC-BY-NC-ND 3.0 license.


\subsection{Fake Video Generation}
All fake face images are sourced from FFHQ~\cite{karras2019style}, with 2{,}964 reference images selected to match the demographic distribution of the real subset. For reference audio, we use speech segments from the TED-LIUM corpus (LIUM)~\cite{rousseau2014enhancing}, a large-scale English speech dataset.\footnote{\url{https://huggingface.co/datasets/kfajdsl/tedlium/}} All audio clips are trimmed to a maximum duration of 60 seconds. Image--audio pairings are established via random matching, ensuring no semantic or identity-level correspondence exists between the visual and auditory streams. The reference-image and driving-audio identities are therefore not intentionally matched in the standard synthesis pipeline.                                                                                                                            
  \begin{enumerate}[leftmargin=*]
      \item \textbf{AniPortrait~\cite{wei2024aniportrait}:} AniPortrait is a diffusion-based framework for facial animation synthesis. It generates high-quality, realistic, and temporally coherent facial videos from input. The method decomposes the process into two stages: the first stage converts extracted audio features into a sequence of 2D facial landmarks to capture fine-grained lip movements and expressions; the second stage employs a diffusion model to synthesize high-fidelity facial videos conditioned on these landmarks. This method supports both audio-driven and video-driven modes; however, we only employ the audio-driven mode. The code repository is available at: \url{https://github.com/Zejun-Yang/AniPortrait}. The license is Apache-2.0 license.
      \item \textbf{Ditto~\cite{li2025ditto}:} Ditto is a diffusion-based talking head framework that enables fine-grained control and real-time inference. The method first extracts motion representations disentangled from identity in the source images. It then leverages a Diffusion Transformer (DiT) to predict audio-synchronized facial and head motion sequences in the motion space. Finally, a face renderer sequentially synthesizes the target image sequence from the predicted motion sequences. This framework enables streaming processing and has fast inference speed. The code repository is available at: \url{https://github.com/antgroup/ditto-talkinghead}. The license is Apache-2.0 license.
      \item \textbf{EDTalk~\cite{tan2024edtalk}:} EDTalk is an Efficient Disentanglement framework for talking head generation. It decomposes complex facial dynamics into three disentangled latent spaces: lip movements, head poses, and emotional expressions. The framework utilizes a set of learnable bases to represent each space and enforces orthogonality among them to ensure the independence of different motion dimensions. Finally, an audio-to-motion module generates dynamic talking videos that are highly consistent with the audio emotion. The code repository is available at: \url{https://github.com/tanshuai0219/EDTalk}. The license is Apache-2.0 license.  
      \item \textbf{Float~\cite{ki2025float}:} FLOAT is an audio-driven talking head video generation method based on the Flow Matching generative model. Its core principle involves shifting the generation process from the traditional pixel space to an efficient and compact Motion Latent Space. By leveraging a Transformer-based vector field predictor with an effective frame-wise conditioning mechanism, FLOAT generates dynamic sequences that are highly synchronized with audio and temporally consistent. These innovations enable the method to overcome the bottlenecks of temporal consistency and inference efficiency in video generation. The code repository is available at: \url{https://github.com/deepbrainai-research/float}. The license is CC BY-NC-ND 4.0 license.  
      \item \textbf{Hallo~\cite{xu2024hallo}:} Unlike traditional techniques that focus solely on lip-sync, the core innovation of Hallo lies in its hierarchical driving architecture. It decomposes the complex task of facial dynamics generation into three levels: lip, expression, and pose. This allows the model to more precisely capture the subtle correlations between audio and facial movements. Adopting a UNet-based diffusion framework, Hallo achieves end-to-end audio-driven visual synthesis. This hierarchical approach also provides adaptive control over expression and pose diversity, enabling more effective personalized generation for different identities. The code repository is available at: \url{https://github.com/fudan-generative-vision/hallo}. The license is MIT license.
      \item \textbf{JoyVASA~\cite{cao2024joyvasa}:} JoyVASA leverages diffusion models to generate facial dynamics and head movements synchronized with audio signals. In the first stage, it employs a disentangled face representation framework to separate dynamic facial expressions from static 3D face representations. In the second stage, a Diffusion Transformer is trained to generate motion sequences directly from audio prompts, independent of character identity. Through this disentangled representation and identity-agnostic motion generation process, JoyVASA extends beyond human portraits to animal faces, demonstrating broad application potential in multilingual support and cross-species animation. The code repository is available at: \url{https://github.com/jdh-algo/JoyVasa}. The license is MIT license. 
      \item \textbf{SadTalker~\cite{zhang2023sadtalker}:} SadTalker employs a 3DMM as its core representation. It learns to generate 3D motion coefficients from audio and utilizes a novel 3D face renderer to synthesize head movements. The model explicitly models the correlations between audio and different types of motion coefficients separately: it uses a custom ExpNet module to extract features from audio and predict facial expression coefficients that are precisely synchronized with speech content; and employs a CVAE-based PoseVAE to generate diverse head motion trajectories. The code repository is available at: \url{https://github.com/OpenTalker/SadTalker}. The license is Apache 2.0 license. 
      \item \textbf{ Sonic~\cite{ji2025sonic}:} Sonic shifts the focus to global audio perception in portrait animation. The method decomposes this perception into intra-clip and inter-clip audio perception to effectively leverage global audio knowledge. For intra-clip perception, Context-enhanced Audio Learning and a Motion-decoupled Controller are employed to extract facial expression information and decouple head motion. For inter-clip perception, Time-aware Position Shift Fusion integrates global information across clips to enable long-audio inference. Ultimately, these two aspects are combined to enhance the overall perception. The code repository is available at: \url{https://github.com/jixiaozhong/Sonic}. The license is CC BY-NC-SA 4.0 license. 
  \end{enumerate}

\subsection{Annotation Quality Control}
\label{sec:annotation}
We apply the FairFace~\cite{karkkainen2021fairface} algorithm to predict demographic attributes (race, gender, and age group) for all candidate videos and images. To mitigate the known error rate of FairFace and ensure label accuracy, we recruit a team of 8 human annotators and implement a rigorous verification pipeline consisting of three stages:

\begin{enumerate}[leftmargin=*]
      \item \textbf{Training.} Each annotator undergoes individual training using diverse examples, with an emphasis on edge cases where FairFace is prone to errors (e.g., ambiguous skin tones and borderline age groups) to ensure consistent labeling standards.
      \item \textbf{Calibration.} Annotators iteratively verify small subsets of FairFace predictions, receive feedback, and repeat until their accuracy exceeds 90\%.
      \item \textbf{Formal Verification.} The selected 2{,}964 entries are randomly distributed among annotators for manual correction of pseudo-label noise.
\end{enumerate}

\noindent All verified annotations undergo expert review under a strict batch auditing protocol. Annotations are evaluated in batches of 100 entries, with 30\% randomly sampled for cross-checking by an expert inspector. A batch is accepted only if disagreements with expert judgment do not exceed three (sampled error rate $\leq$10\%); otherwise, the entire batch is returned for re-annotation and the annotator undergoes retraining. All erroneous samples identified during sampling are individually corrected by the expert.

\subsection{Video Quality Curation}
\label{sec:video_quality_audit}

In our benchmark construction, we recruited three independent reviewers to perform quality screening.
Three independent reviewers screen the generated videos for severe synthesis failures. Videos were discarded only when they exhibited severe generation failures that significantly affected visual realism. The filtering criteria are listed below:
\begin{enumerate}
    \item Only mouth motion is generated while other facial regions remain unchanged;
    \item Strong cross-frame inconsistency between the generated person and the background;
    \item Severe temporal instability, such as noticeable frame jitter or abrupt brightness changes across frames.
\end{enumerate}

To evaluate the consistency among reviewers, we calculated Fleiss' Kappa, achieving a value of $\kappa = 0.896$. Since $\kappa > 0.8$ indicates excellent agreement, this demonstrates strong consistency among the three reviewers.

Furthermore, we performed a quantitative analysis in Table~\ref{tab:curation_quality} between discarded and retained videos. The discarded videos have an average FVD of 0.0599, while the retained videos achieve a lower FVD of 0.0265, indicating better visual quality. Meanwhile, the SyncNet distance is 0.21 for discarded videos and 0.15 for retained videos, suggesting that the retained videos exhibit better audio-visual synchronization.

\begin{table}[htbp]
\centering
\caption{Quality comparison before and after manual selection. Lower values are better. FVD is reported on the normalized scale used in the quality audit.}
\label{tab:curation_quality}
\begin{tabular}{lcc}
\toprule
Collection & Normalized FVD $\downarrow$ & SyncNet distance $\downarrow$ \\
\midrule
Discarded & 0.0599 & 0.21 \\
Retained & 0.0265 & 0.15 \\
\bottomrule
\end{tabular}
\end{table}

\subsection{Train/Validation/Test Splits and Identity Separation}
We enforce strict identity isolation across all data splits. The training, validation, and test sets (split at an 8:1:1 ratio) contain mutually exclusive sets of identities, ensuring that no individual appears in more than one split. This prevents direct identity overlap between the training, validation, and testing splits. The standard benchmark also uses disjoint real and fake identity pools; Protocol~4 separately evaluates identity-paired synthesis.

\subsection{Demographic Distribution of the Balanced Subset}
\label{sec:balanced_distribution}                           
                                   
\begin{figure}[t]
      \centering   
      \includegraphics[width=0.9\linewidth]{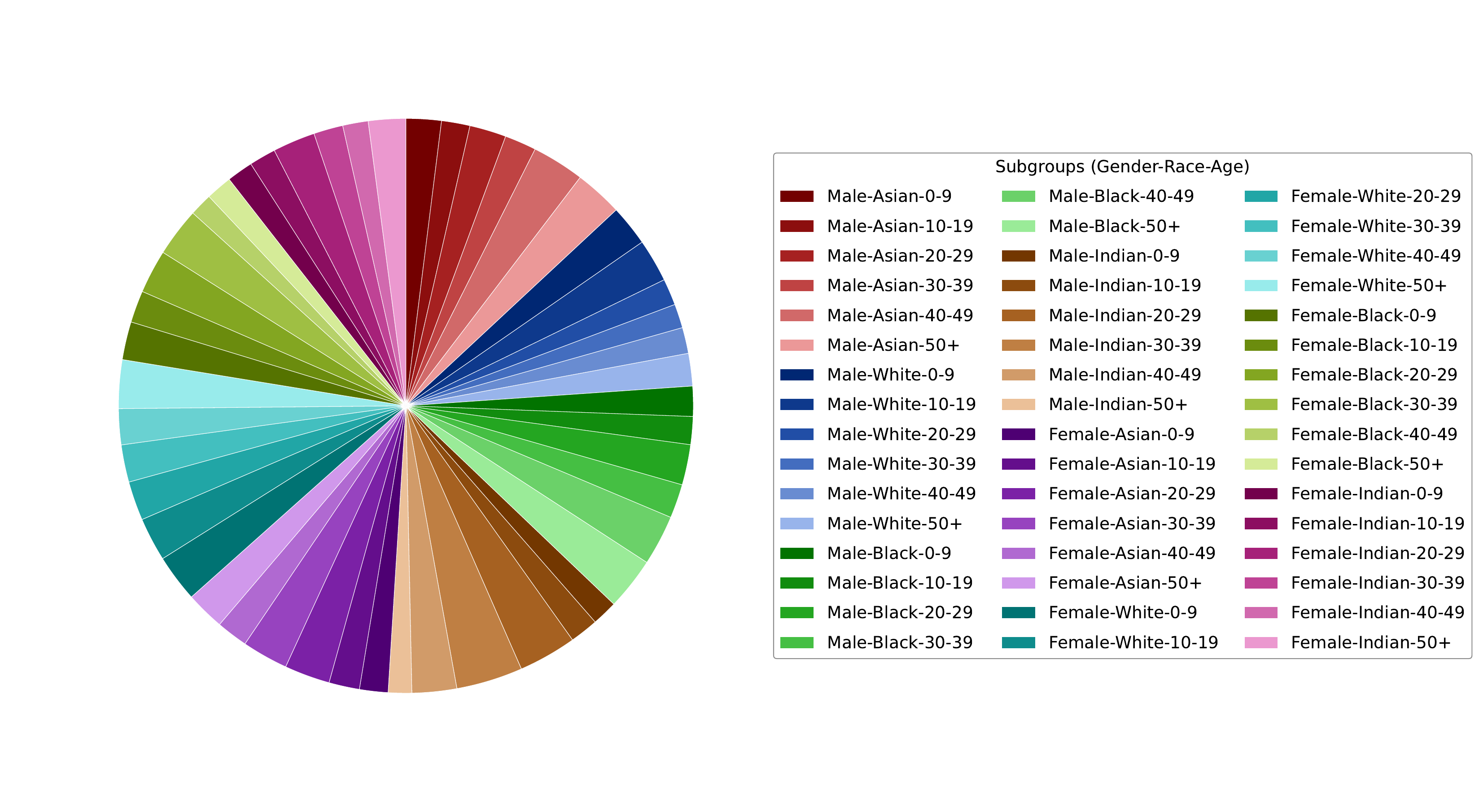}
      \vspace{-19pt}
      \caption{Distribution of the 48 intersectional demographic subgroups in the balanced subset. All subgroups are approximately equal in size, confirming no long-tail effect across joint demographic attributes.}
      \label{fig:balanced_48subgroups}
  \end{figure}

As shown in Figure~\ref{fig:balanced_48subgroups}, the 48 intersectional subgroups (2 genders $\times$ 4 races $\times$ 6 age groups) in our balanced subset are approximately equal in size with no severe long-tail effect, confirming balanced coverage across joint demographic attributes and train/val/test splits.                                                      

\subsection{Computational Resources}
 Generating FairTalking-Bench requires approximately 25,000 NVIDIA V100 GPU-hours. We release all generated videos so that downstream users can use the benchmark without repeating this computationally intensive generation process, making it more accessible to research groups with limited computational resources.

\section{Data Quality and Source Distribution Analysis}

\subsection{Quality Confounder Analysis}
\label{sec:confounder}

We examine demographic performance differences and selected low-level quality statistics within each data source.

\subsubsection{Performance Disparities under Identical Configurations}
\begin{figure*}[t]
    \centering
    \includegraphics[width=\textwidth]{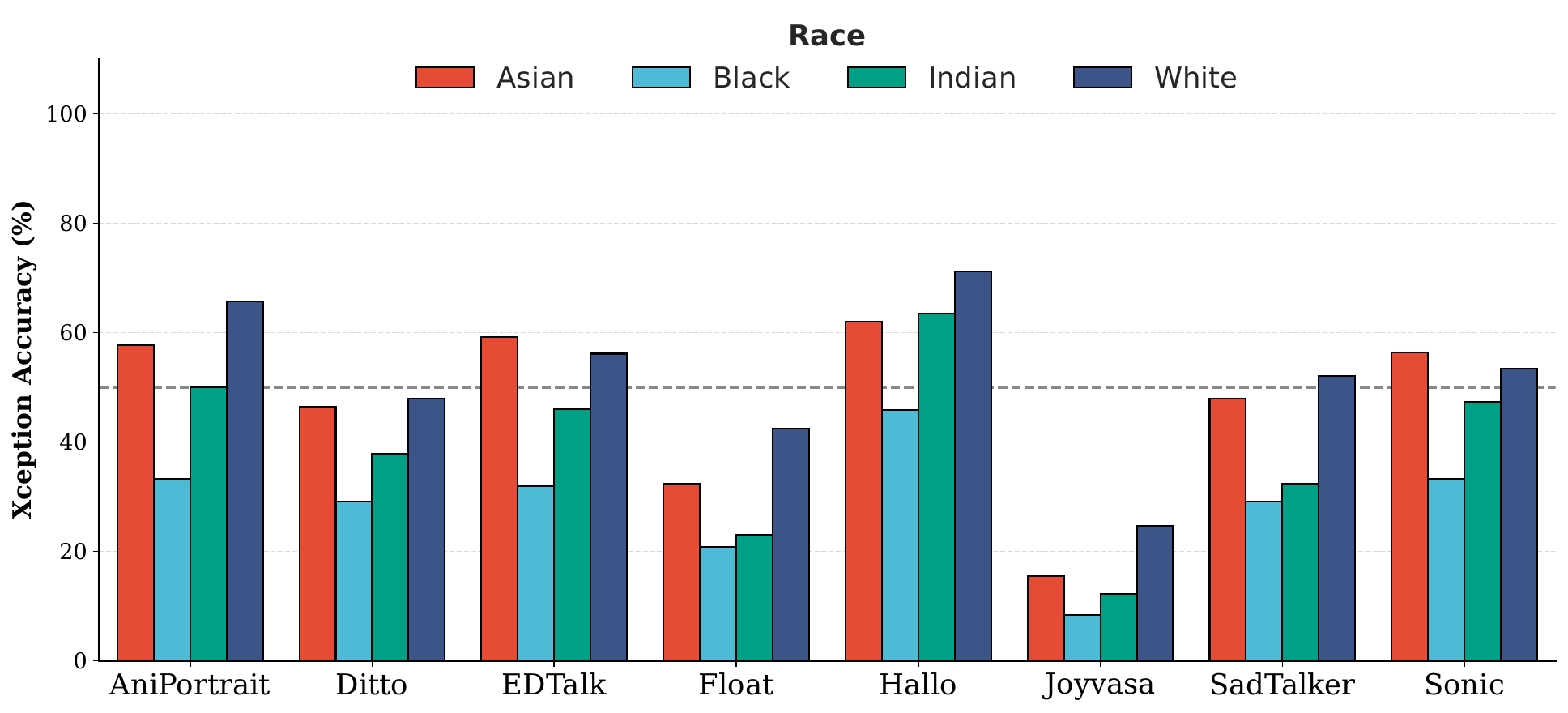}
    \caption{Detection accuracy across four racial groups under identical generative configurations. Performance differs across the evaluated racial groups and generators.}
    \label{fig:quality_confounder}
\end{figure*}

As shown in Figure~\ref{fig:quality_confounder}, we evaluate detectors under identical generative configurations across demographics. We observe that the Black race consistently achieves the lowest detection accuracy across nearly all eight generators, while the White race consistently achieves the highest. The generator configuration is fixed within each comparison, so the plotted differences describe demographic performance variation under the same synthesis configuration.

\subsubsection{Quality Parity across Demographics}
We examine brightness and contrast in the underlying real videos using the following preprocessing and within-source tests:

\begin{itemize}[leftmargin=*]
    \item \textbf{Data Preprocessing:} We applied a global normalization protocol to the luminance and dynamic range of frames across all sources. This aligns the measured luminance and dynamic-range statistics across recording sources.
    
    \item \textbf{Outlier Removal:} Extreme recording artifacts (top and bottom 5\% of the quality distribution) were excluded within each demographic group to ensure the statistical analysis reflects the general characteristics of the dataset.
    
    \item \textbf{Statistical Validation:} We utilized Kruskal-Wallis (KW) tests to assess group-level disparities. As summarized in Table~\ref{tab:kw_results}, the tests for \textbf{brightness} and \textbf{contrast} yielded $p$-values consistently above the 0.05 threshold across all sources. 
\end{itemize}

\textbf{Results:} The within-source tests do not detect racial-group differences in the measured brightness and contrast statistics at the 0.05 level. These results characterize the two selected quality measures; source-matched and region-level analyses are provided in Section~\ref{sec:source_audit}.

\begin{table}[t]
    \centering
    \caption{Kruskal--Wallis tests across racial groups within each
    data source. All reported $p$-values exceed 0.05 for the two
    measured statistics.}
    \label{tab:kw_results}
    \scriptsize
    \setlength{\tabcolsep}{2pt}
    \renewcommand{\arraystretch}{1.05}
    \begin{tabular*}{\linewidth}{@{\extracolsep{\fill}}llccc@{}}
        \toprule
        \textbf{Source} & \textbf{Metric}
        & \shortstack{\textbf{H-}\\\textbf{statistic}}
        & \textbf{$p$-value}
        & \shortstack{\textbf{Significant?}\\\textbf{($p<0.05$)}} \\
        \midrule
        CelebV-HQ & Brightness & 2.6680 & 0.1024 & False \\
        CelebV-HQ & Contrast   & 3.6049 & 0.1649 & False \\
        \midrule
        DFDC & Brightness & 0.7265 & 0.6954 & False \\
        DFDC & Contrast   & 1.1221 & 0.5706 & False \\
        \midrule
        HDTF & Brightness & 0.0408 & 0.9798 & False \\
        HDTF & Contrast   & 4.1363 & 0.1264 & False \\
        \bottomrule
    \end{tabular*}
\end{table}

\subsection{Source-Bias Analysis}
\label{sec:source_bias_analysis}

We agree that source bias is an important concern when real and fake samples are collected from different sources. To investigate whether the detector relies on source-specific characteristics rather than forgery artifacts, we conduct additional analyses.

For real videos from different sources (DFDC, CelebV-HQ, and HDTF), we first compute the cosine similarity of their frequency-domain features. As shown in Table~\ref{tab:source_frequency_similarity}, the similarities among the three real sources are consistently high (0.99--1.00), indicating that these datasets share similar frequency characteristics.

\begin{table}[t]
    \centering
    \caption{Cosine similarity of frequency-domain features across real-video sources.}
    \label{tab:source_frequency_similarity}
    \scriptsize
    \setlength{\tabcolsep}{4pt}
    \renewcommand{\arraystretch}{1.05}
    \begin{tabular*}{\linewidth}{@{\extracolsep{\fill}}lccc@{}}
        \toprule
        \textbf{Real source} & \textbf{DFDC}
        & \textbf{CelebV-HQ} & \textbf{HDTF} \\
        \midrule
        DFDC     & 1      & 0.9984 & 0.9934 \\
        CelebV-HQ & 0.9984 & 1      & 0.9981 \\
        HDTF     & 0.9934 & 0.9981 & 1      \\
        \bottomrule
    \end{tabular*}
\end{table}

Furthermore, we perform a cross-source evaluation by training the detector using real videos from one source combined with fake videos, and testing it on real videos from the other two unseen sources. Table~\ref{tab:cross_source_evaluation} shows that Xception maintains ACC values between 97.04\% and 98.62\% across the six source combinations, demonstrating stable performance when the real-video source changes.

\begin{table}[t]
    \centering
    \caption{Cross-source evaluation of Xception with different real-video sources.}
    \label{tab:cross_source_evaluation}
    \scriptsize
    \setlength{\tabcolsep}{3pt}
    \renewcommand{\arraystretch}{1.05}
    \begin{tabular*}{\linewidth}{@{\extracolsep{\fill}}lllr@{}}
        \toprule
        \textbf{Model}
        & \shortstack{\textbf{Training}\\\textbf{real source}}
        & \shortstack{\textbf{Testing}\\\textbf{real source}}
        & \textbf{ACC (\%)} \\
        \midrule
        \multirow{6}{*}{Xception}
        & \multirow{2}{*}{DFDC}
        & CelebV-HQ & 97.35 \\
        & & HDTF & 98.10 \\
        \cmidrule(lr){2-4}
        & \multirow{2}{*}{CelebV-HQ}
        & DFDC & 98.44 \\
        & & HDTF & 98.62 \\
        \cmidrule(lr){2-4}
        & \multirow{2}{*}{HDTF}
        & DFDC & 97.61 \\
        & & CelebV-HQ & 97.04 \\
        \bottomrule
    \end{tabular*}
\end{table}\subsection{Source-Matched and Region-Level Distribution Analysis}
\label{sec:source_audit}
\subsubsection{Source-Matched Construction and Sampling}
The standard benchmark combines real videos from DFDC, CelebV-HQ, and HDTF with synthetic videos generated from FFHQ images and TED-LIUM audio. We construct an HDTF-matched control by synthesizing videos from reference frames and driving audio taken from the corresponding real HDTF videos. This matches the input identity, reference-frame source, background source, and driving audio. The matched control is implemented on HDTF; it is not a source-matched reconstruction of all three real-source collections.

For each setting, the distributional audit samples 400 real and 400 fake videos and repeats sampling 50 times. We report the mean and standard deviation of normalized Fr\'echet Video Distance (FVD). For audio, tracks are resampled to 16\,kHz mono, embedded with pretrained VGGish, and summarized using Fr\'echet Audio Distance (FAD). FAD uses the same sample-count and repeated-sampling protocol; the table reports its point estimate.

\begin{table*}[htbp]
\centering
\caption{Full-frame video and audio distribution distances. FVD values are normalized. FF++ has no audio track in this comparison, so FAD is not applicable.}
\label{tab:source_distances}
\begin{tabular}{llrrr}
\toprule
Setting & Real source & FVD mean & FVD std & FAD \\
\midrule
FairTalking-Bench & CelebV-HQ & 0.02550 & 0.00300 & 0.4653 \\
FairTalking-Bench & DFDC & 0.02781 & 0.00139 & 0.4415 \\
FairTalking-Bench & HDTF & 0.02618 & 0.00798 & 0.4229 \\
HDTF-matched & HDTF & 0.02505 & 0.00142 & 0.3323 \\
TalkingHeadBench & CelebV-HQ & 0.13408 & 0.03765 & 0.5237 \\
TalkingHeadBench & FF++ & 0.27875 & 0.07450 & -- \\
\bottomrule
\end{tabular}
\end{table*}

The HDTF-matched setting has a full-frame FVD of 0.02505, compared with 0.02618 for the standard HDTF real-source comparison, and reduces FAD from 0.4229 to 0.3323. The normalized FVDs for the evaluated TalkingHeadBench comparisons are higher. These results characterize distributional differences in the selected video and audio embedding spaces.

\subsubsection{Face and Background Decomposition}
We construct face masks using dlib's 68-point facial landmarks and compute FVD separately for the masked face region and its complementary background region, using the same 400-real/400-fake and 50-resampling protocol.

\begin{table}[htbp]
\centering
\caption{Region-level normalized FVD. Face and background FVDs are computed independently and are not additive components of full-frame FVD.}
\label{tab:region_distances}
\begin{tabular}{lrr}
\toprule
Setting & Face FVD & Background FVD \\
\midrule
FairTalking-Bench & 0.017825 & 0.00935 \\
HDTF-matched & 0.004860 & 0.00907 \\
TalkingHeadBench & 0.060600 & 0.01520 \\
\bottomrule
\end{tabular}
\end{table}

Source matching reduces face-region FVD from 0.017825 to 0.004860, while background-region FVD changes from 0.00935 to 0.00907. Thus, the measured change is concentrated in the facial embedding distribution. Face-region FVD includes effects of appearance, pose, expression, identity, and synthesis quality; these distributional measurements complement the detection experiments rather than directly measuring which cues determine a detector's decision.

\section{Supplementary Demographic Fairness Evaluation}
\label{sec:extended_fairness}
\subsection{Fairness Metrics Details}
\label{sec:metrics}
Following the experimental setting in recent research~\cite{lin2024preserving}, we adopt four fairness metrics to evaluate the bias of Deepfake detectors. Let $\mathcal{D} = \{(X_i, Y_i, A_i)\}_{i=1}^N$ denote the test set, where $X_i$ is the input image, $Y_i \in \{0, 1\}$ is the ground truth label ($0$ for real, $1$ for fake), and $A_i \in \mathcal{G}$ represents the demographic attribute (subgroup), where $\mathcal{G}$ is the set of all subgroups. Let $\hat{Y}_i$ denote the predicted label. The indicator function is denoted by $\mathbb{I}[\cdot]$.

\vspace{5pt}
\noindent \textbf{1. False Positive Rate Disparity ($F_{\text{FPR}}$)}: 
This metric measures the disparity in False Positive Rate (FPR) across different demographic groups compared to the overall population. It is calculated as the sum of the absolute difference between each subgroup's FPR and the global FPR:
\begin{equation}
    F_{\text{FPR}} = \sum_{g \in \mathcal{G}} \left| \text{FPR}_g - \text{FPR}_{\text{overall}} \right|,
\end{equation}
where $\text{FPR}_g = \frac{\sum_{i=1}^N \mathbb{I}[\hat{Y}_i=1, Y_i=0, A_i=g]}{\sum_{i=1}^N \mathbb{I}[Y_i=0, A_i=g]}$.

\vspace{5pt}
\noindent \textbf{2. Overall Accuracy Equality ($F_{\text{OAE}}$)}: 
This metric captures the maximum disparity in Accuracy (ACC) among all demographic groups. It is defined as the difference between the maximum and minimum group accuracy:
\begin{equation}
    F_{\text{OAE}} = \max_{g \in \mathcal{G}} (\text{ACC}_g) - \min_{g' \in \mathcal{G}} (\text{ACC}_{g'}),
\end{equation}
where $\text{ACC}_g = \frac{\sum_{i=1}^N \mathbb{I}[\hat{Y}_i=Y_i, A_i=g]}{\sum_{i=1}^N \mathbb{I}[A_i=g]}$.

\vspace{5pt}
\noindent \textbf{3. Demographic Parity ($F_{\text{DP}}$)}: 
$F_{\text{DP}}$ measures the maximum difference in the prediction rates (Positive Prediction Rate or Negative Prediction Rate) across all demographic groups. It evaluates whether the model predicts a specific class $k$ (real or fake) at the same rate regardless of the demographic attribute:
\begin{equation}
\begin{aligned}
F_{\text{DP}}
= \max_{k \in \{0,1\}} \Bigl(
    &\max_{g \in \mathcal{G}}
      P(\hat{Y}=k \mid A=g) \\
    &-\min_{g' \in \mathcal{G}}
      P(\hat{Y}=k \mid A=g')
\Bigr).
\end{aligned}
\end{equation}
\vspace{5pt}
\noindent \textbf{4. Equalized Odds ($F_{\text{MEO}}$)}: 
Also known as Max Equalized Opportunity, this metric captures the largest disparity in prediction outcomes (specifically True Positive Rates and True Negative Rates) when comparing different demographic groups. It is formulated as:
\begin{equation}
\begin{aligned}
F_{\text{MEO}}
= \max_{k \in \{0,1\}} \Bigl(
    &\max_{g \in \mathcal{G}}
      P(\hat{Y}=k \mid Y=k, A=g) \\
    &-\min_{g' \in \mathcal{G}}
      P(\hat{Y}=k \mid Y=k, A=g')
\Bigr).
\end{aligned}
\end{equation}
Essentially, it calculates the worst-case disparity in TPR (when $k=1$) and TNR (when $k=0$) across groups. For all the metrics above ($F_{\text{FPR}}$, $F_{\text{OAE}}$, $F_{\text{DP}}$, $F_{\text{MEO}}$), a \textbf{lower value indicates better fairness} performance.

\subsection{Five Subsets with Distinct Distributions}
\label{sec:five_subset}
\begin{figure*}[h]
    \centering
    \includegraphics[width=\textwidth]{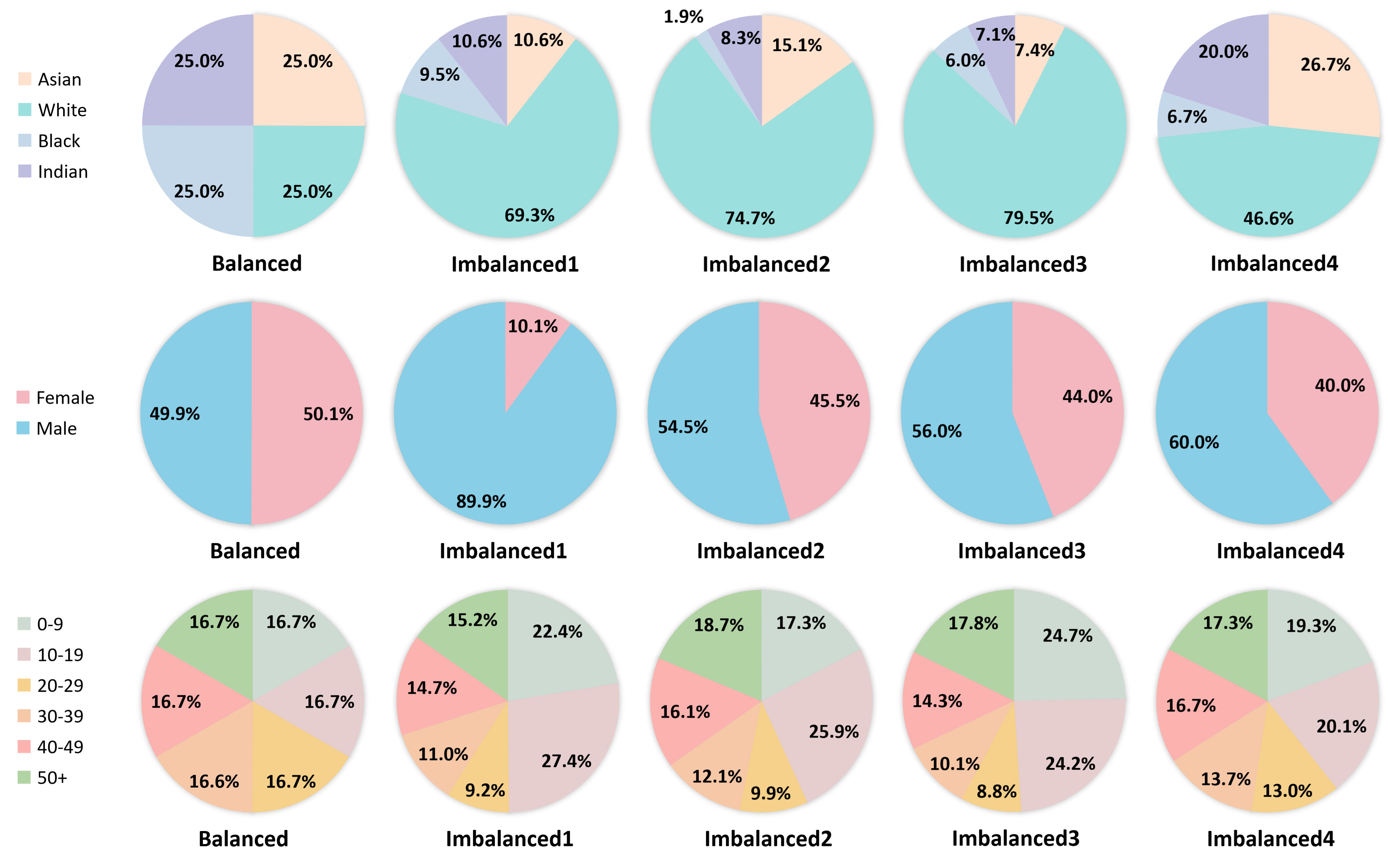}
    \caption{Demographic distributions of the five subsets used in \textbf{Protocol 2}, including one balanced subset and four imbalanced subsets. The distributions are shown across three attributes: race, gender, and age group}
    \label{fig:pipeline1}
    \vspace{-8pt}
\end{figure*}

Specifically, as shown in~\ref{fig:pipeline1} the five subsets are: Balanced (4,743 samples with equal demographic proportions), and four Imbalanced subsets (1,150, 945, 1,489, 4,743 samples) simulating real-world long-tailed demographic scarcity(e.g., majority White/Male/Young).

\subsection{Evaluation with Unequal Training Sizes}
\label{sec:protocol2_unequal}
The original Protocol 2 evaluation compares one balanced training subset and four imbalanced subsets on the same balanced test set. The balanced subset contains 4,743 videos, whereas the imbalanced subsets contain 945--4,743 videos. Table~\ref{tab:metrics_comparison1} preserves these original results. Because both demographic composition and training-set size vary, this experiment characterizes their combined changes and does not isolate the effect of demographic imbalance. The matched-count comparisons in the main text provide the primary evidence for Protocol 2.

\begin{table}[htbp]
    \centering
    \caption{Protocol 2 -- detection and fairness performance on a shared balanced test set. Training subsets differ in both demographic composition and sample count; this comparison does not isolate demographic composition from training-set size.}
    \label{tab:metrics_comparison1}
    \resizebox{\linewidth}{!}{%
    \scriptsize
    \setlength{\tabcolsep}{2.5pt}
    \renewcommand{\arraystretch}{0.9}
    \begin{tabular}{llrrrrrrrr}
        \toprule
        \multirow{3}{*}[-1.5ex]{Training subset}
        & \multirow{3}{*}[-1.5ex]{Detector}
        & \multicolumn{8}{c}{Testing subset (balanced)} \\
        \cmidrule(lr){3-10}
        & & \multicolumn{4}{c}{Detection metrics (\%) $\uparrow$}
        & \multicolumn{4}{c}{Fairness metrics (\%) $\downarrow$} \\
        \cmidrule(lr){3-6}\cmidrule(lr){7-10}
        & & AUC & T1 & T0.1 & ACC
        & $F_{\text{FPR}}$ & $F_{\text{OAE}}$
        & $F_{\text{DP}}$ & $F_{\text{MEO}}$ \\
        \midrule
        \multirow{4}{*}{Balanced}
        & Xception~\cite{rossler2019faceforensics++}
        & 99.98 & 99.53 & 98.53 & 99.35 & 41.05 & 2.47 & 5.63 & 22.22 \\
        & FTCN~\cite{Zheng_2021_ICCV}
        & 99.54 & 90.00 & 37.03 & 98.20 & 63.55 & 3.39 & 5.94 & 30.95 \\
        & TALL~\cite{xu2023tall}
        & 99.92 & 97.11 & 89.99 & 99.50 & 27.33 & 1.24 & 5.63 & 11.11 \\
        & Adapter~\cite{cui2025forensics}
        & 99.99 & 99.91 & 97.75 & 99.77 & 14.34 & 0.62 & 5.63 & 5.56 \\
        \midrule
        \multirow{4}{*}{Imbalanced1}
        & Xception~\cite{rossler2019faceforensics++}
        & 99.94 & 98.02 & 94.27 & 99.16 & 31.80 & 2.16 & 4.85 & 16.67 \\
        & FTCN~\cite{Zheng_2021_ICCV}
        & 98.78 & 69.31 & 7.11 & 97.59 & 55.14 & 3.71 & 7.92 & 31.20 \\
        & TALL~\cite{xu2023tall}
        & 99.68 & 93.44 & 74.92 & 98.89 & 45.15 & 2.48 & 5.94 & 22.22 \\
        & Adapter~\cite{cui2025forensics}
        & 99.26 & 89.24 & 51.25 & 97.47 & 54.50 & 3.40 & 5.34 & 21.15 \\
        \midrule
        \multirow{4}{*}{Imbalanced2}
        & Xception~\cite{rossler2019faceforensics++}
        & 99.96 & 98.53 & 95.52 & 99.54 & 70.83 & 4.33 & 7.63 & 37.82 \\
        & FTCN~\cite{Zheng_2021_ICCV}
        & 98.47 & 55.47 & 7.04 & 97.89 & 70.83 & 4.33 & 7.63 & 37.82 \\
        & TALL~\cite{xu2023tall}
        & 99.57 & 88.26 & 70.18 & 98.24 & 38.47 & 2.50 & 4.82 & 16.67 \\
        & Adapter~\cite{cui2025forensics}
        & 99.97 & 99.42 & 94.31 & 99.60 & 16.84 & 1.23 & 5.33 & 5.56 \\
        \midrule
        \multirow{4}{*}{Imbalanced3}
        & Xception~\cite{rossler2019faceforensics++}
        & 99.91 & 99.05 & 85.04 & 99.50 & 23.60 & 1.23 & 4.86 & 8.33 \\
        & FTCN~\cite{Zheng_2021_ICCV}
        & 98.91 & 71.25 & 5.56 & 98.05 & 56.68 & 3.10 & 6.08 & 30.95 \\
        & TALL~\cite{xu2023tall}
        & 99.57 & 89.94 & 52.57 & 98.93 & 40.69 & 2.81 & 5.33 & 22.22 \\
        & Adapter~\cite{cui2025forensics}
        & 99.93 & 99.25 & 88.98 & 99.49 & 11.00 & 0.90 & 6.23 & 3.03 \\
        \midrule
        \multirow{4}{*}{Imbalanced4}
        & Xception~\cite{rossler2019faceforensics++}
        & 99.50 & 97.52 & 87.89 & 99.06 & 51.11 & 2.46 & 7.14 & 29.17 \\
        & FTCN~\cite{Zheng_2021_ICCV}
        & 99.00 & 75.45 & 29.44 & 97.20 & 77.19 & 3.05 & 6.35 & 49.46 \\
        & TALL~\cite{xu2023tall}
        & 98.97 & 88.59 & 67.95 & 97.43 & 49.38 & 3.53 & 5.75 & 22.22 \\
        & Adapter~\cite{cui2025forensics}
        & 99.72 & 95.52 & 79.88 & 96.54 & 29.80 & 1.78 & 5.64 & 5.67 \\
        \bottomrule
    \end{tabular}%
    }
\end{table}

\subsection{Evaluation at a Smaller Matched Training Size}
\label{sec:protocol2_small}
We additionally construct balanced and imbalanced subsets with matched sample counts (1,104 each), identity counts, real/fake ratios, and generator distributions — the only remaining difference is demographic composition (imbalanced: male 65.3\%, White 39.7\% vs. Black 11.4\%, age 30–39 27.3\% vs. 40–49 10.1\%). As Table~\ref{tab:matched_1104} shows, detectors trained with balanced demographic distributions consistently achieve better detection performance and lower fairness disparities compared with those trained on imbalanced subsets. This indicates that the robustness improvement is not merely caused by a larger training set size, but is closely related to the demographic diversity and balance of the training distribution. Therefore, demographic balance is an important factor for improving both generalization and fairness in deepfake detection.

\begin{table}[t]
\centering
\caption{Matched-size comparison with 1,104 real and 1,104 fake training videos per condition. Values are percentages; detection metrics are higher-is-better, and disparity metrics are lower-is-better.}
\label{tab:matched_1104}
\resizebox{\linewidth}{!}{%
\begin{tabular}{llrrrrrrrr}
\toprule
Condition & Detector & AUC & T1 & T0.1 & ACC & $F_{\mathrm{FPR}}$ & $F_{\mathrm{OAE}}$ & $F_{\mathrm{DP}}$ & $F_{\mathrm{MEO}}$ \\
\midrule
Balanced & Xception & 98.99 & 96.40 & 86.89 & 98.48 & 21.92 & 1.49 & 6.44 & 25.11 \\
Balanced & FTCN & 98.76 & 65.85 & 28.44 & 97.76 & 64.84 & 2.17 & 6.14 & 13.58 \\
Balanced & TALL & 98.97 & 81.16 & 66.95 & 96.21 & 21.29 & 1.23 & 6.73 & 11.11 \\
Balanced & Adapter & 98.66 & 94.93 & 78.88 & 98.65 & 16.45 & 1.16 & 6.56 & 19.44 \\
\midrule
Imbalanced & Xception & 97.90 & 88.52 & 66.67 & 98.16 & 60.18 & 3.91 & 8.65 & 39.17 \\
Imbalanced & FTCN & 98.55 & 48.26 & 18.63 & 96.50 & 87.55 & 4.16 & 6.35 & 35.44 \\
Imbalanced & TALL & 98.07 & 78.69 & 57.92 & 94.66 & 41.67 & 4.39 & 7.49 & 17.47 \\
Imbalanced & Adapter & 98.27 & 86.50 & 62.28 & 97.27 & 36.80 & 2.51 & 7.75 & 26.67 \\
\bottomrule
\end{tabular}}
\end{table}

\subsection{Statistical Stability of Fairness Metrics}
\label{sec:statistical_stability}

To validate the statistical stability of our fairness evaluation, we perform repeated resampling of the test set for each imbalance severity in Protocol 2 and report the mean and standard deviation of fairness metrics across runs using the Xception~\cite{rossler2019faceforensics++} detector. Table~\ref{tab:stability} reports the observed sampling variability for these original unequal-size subsets. The dispersion differs across metrics and training subsets; these summaries are not confidence intervals for the matched-size comparison in the main text.

\begin{table}[t]
    \centering
    \caption{Statistical stability of fairness metrics under repeated
    resampling (Xception). We report mean $\pm$ standard deviation
    across multiple runs for each imbalance severity in Protocol 2.}
    \label{tab:stability}
    \scriptsize
    \setlength{\tabcolsep}{1.5pt}
    \renewcommand{\arraystretch}{1.05}
    \begin{tabular}{@{}lcccc@{}}
        \toprule
        \textbf{Split}
        & $F_{\text{FPR}}$
        & $F_{\text{OAE}}$
        & $F_{\text{DP}}$
        & $F_{\text{MEO}}$ \\
        \midrule
        Imbalanced (945)
        & 42.00 $\pm$ 8.50
        & 2.07 $\pm$ 2.00
        & \textbf{4.83 $\pm$ 2.20}
        & 16.10 $\pm$ 3.80 \\
        Imbalanced (1150)
        & 82.94 $\pm$ 6.10
        & 19.00 $\pm$ 2.80
        & 18.50 $\pm$ 2.90
        & 21.10 $\pm$ 5.00 \\
        Imbalanced (1489)
        & 39.70 $\pm$ 9.90
        & 7.66 $\pm$ 1.30
        & 10.10 $\pm$ 2.10
        & 15.40 $\pm$ 4.50 \\
        Imbalanced (4743)
        &35.90 $\pm$ 7.10
        &8.53 $\pm$ 2.40
        & 9.85 $\pm$ 2.00
        & 18.08 $\pm$ 4.30 \\
        \midrule
        Balanced (4743)
        & \textbf{26.01 $\pm$ 4.90}
        & \textbf{0.60 $\pm$ 3.20}
        & 5.30 $\pm$ 3.40
        & \textbf{11.50 $\pm$ 3.50} \\
        \bottomrule
    \end{tabular}
\end{table}

\subsection{Grouping and Bootstrap Intervals}
The five-subset fairness results use the eight gender-by-race intersections. Table~\ref{tab:balanced_fairness_ci} supplements the balanced training condition with 95\% bootstrap intervals. The intervals describe uncertainty in the reported test-set estimates, rather than variability across independently trained models.

\begin{table}[htbp]
\centering
\caption{Fairness estimates and 95\% bootstrap intervals for the balanced 4,743-video training condition, evaluated over the eight gender-by-race subgroups. Values are percentage-scaled.}
\label{tab:balanced_fairness_ci}
\resizebox{\linewidth}{!}{%
\begin{tabular}{lrrrr}
\toprule
Detector & $F_{\mathrm{FPR}}$ & $F_{\mathrm{OAE}}$ & $F_{\mathrm{DP}}$ & $F_{\mathrm{MEO}}$ \\
\midrule
Xception & 41.05 [24.75, 69.06] & 2.47 [1.19, 4.10] & 5.63 [4.05, 11.58] & 22.22 [10.26, 36.00] \\
FTCN & 63.55 [44.48, 103.46] & 3.39 [2.29, 5.72] & 5.94 [4.37, 11.80] & 30.95 [21.12, 68.72] \\
TALL & 27.33 [12.47, 50.14] & 1.24 [0.67, 2.74] & 5.63 [4.08, 11.06] & 11.11 [5.00, 23.53] \\
Adapter & 14.34 [4.85, 34.22] & 0.62 [0.30, 1.57] & 5.63 [4.13, 11.57] & 5.56 [2.70, 14.71] \\
\bottomrule
\end{tabular}}
\end{table}

\subsection{Full Three-Way Intersectional Evaluation}
We additionally evaluate fairness over all 48 gender-by-race-by-age subgroups. Tables \label{tab:fairness_48_bootstrap_ci}report the results for the five training subsets. The finer grouping exposes substantial variation: for balanced training, Adapter obtains $F_{\mathrm{MEO}}=9.09\%$, compared with 33.33\% for TALL and 100.00\% for Xception and FTCN. The wider intervals quantify the uncertainty associated with the fine-grained comparisons. Since $F_{\mathrm{FPR}}$ sums deviations over groups, its percentage-scaled value can exceed 100 and depends on the number of groups; values for eight and 48 groups should not be interpreted as directly comparable normalized scores.

\begin{table*}[t]
    \centering
    \caption{Fairness metrics with 95\% bootstrap confidence intervals.
    Each entry reports the point estimate followed by its confidence interval.}
    \label{tab:fairness_bootstrap_ci}
    \scriptsize
    \setlength{\tabcolsep}{4pt}
    \renewcommand{\arraystretch}{1.05}
    \begin{tabular*}{\textwidth}{@{\extracolsep{\fill}}llcccc@{}}
        \toprule
        \textbf{Training} & \textbf{Detector}
        & $F_{\text{FPR}}$ (CI)
        & $F_{\text{OAE}}$ (CI)
        & $F_{\text{DP}}$ (CI)
        & $F_{\text{MEO}}$ (CI) \\
        \midrule
        \multirow{4}{*}{Balanced (4,743)}
        & Xception & 41.05 [24.75, 69.06] & 2.47 [1.19, 4.10] & 5.63 [4.05, 11.58] & 22.22 [10.26, 36.00] \\
        & FTCN & 63.55 [44.48, 103.46] & 3.39 [2.29, 5.72] & 5.94 [4.37, 11.80] & 30.95 [21.12, 68.72] \\
        & TALL & 27.33 [12.47, 50.14] & 1.24 [0.67, 2.74] & 5.63 [4.08, 11.06] & 11.11 [5.00, 23.53] \\
        & FA & 14.34 [4.85, 34.22] & 0.62 [0.30, 1.57] & 5.63 [4.13, 11.57] & 5.56 [2.70, 14.71] \\
        \midrule
        \multirow{4}{*}{Imbalanced1 (945)}
        & Xception & 31.80 [23.33, 59.88] & 2.16 [1.22, 3.76] & 4.85 [3.72, 11.12] & 16.67 [10.00, 29.17] \\
        & FTCN & 55.14 [40.89, 94.39] & 3.71 [2.60, 6.16] & 7.92 [5.19, 12.94] & 31.20 [19.85, 51.36] \\
        & TALL & 45.15 [26.49, 74.19] & 2.48 [1.65, 4.48] & 5.94 [4.06, 11.65] & 22.22 [10.87, 36.37] \\
        & FA & 54.50 [34.19, 87.93] & 4.32 [2.33, 5.75] & 5.34 [3.72, 10.59] & 24.15 [15.55, 38.59] \\
        \midrule
        \multirow{4}{*}{Imbalanced2 (1,150)}
        & Xception & 67.10 [50.71, 108.08] & 2.59 [2.74, 6.75] & 6.33 [5.18, 12.70] & 33.00 [23.83, 57.15] \\
        & FTCN & 70.83 [43.44, 115.98] & 4.33 [1.95, 7.48] & 7.63 [5.92, 10.65] & 37.82 [26.49, 49.34] \\
        & TALL & 38.47 [22.97, 64.50] & 2.50 [2.00, 4.50] & 4.82 [4.00, 10.97] & 16.67 [9.37, 29.42] \\
        & FA & 16.84 [7.82, 36.07] & 1.23 [0.61, 2.64] & 5.33 [3.91, 11.28] & 5.56 [2.94, 15.38] \\
        \midrule
        \multirow{4}{*}{Imbalanced3 (1,489)}
        & Xception & 23.65 [12.84, 45.30] & 1.23 [0.65, 2.66] & 4.86 [3.58, 11.03] & 8.33 [5.00, 19.44] \\
        & FTCN & 56.68 [40.34, 92.24] & 3.10 [2.20, 5.41] & 6.68 [4.38, 11.92] & 30.95 [19.17, 46.88] \\
        & TALL & 40.69 [25.67, 71.49] & 2.81 [2.11, 5.03] & 5.33 [4.10, 11.20] & 22.22 [10.53, 36.37] \\
        & FA & 11.00 [4.35, 27.21] & 0.90 [0.60, 2.19] & 6.23 [4.09, 11.71] & 3.03 [2.27, 11.77] \\
        \midrule
        \multirow{4}{*}{Imbalanced4 (4,743)}
        & Xception & 53.05 [39.59, 69.85] & 2.63 [0.98, 5.99] & 6.75 [5.80, 13.31] & 29.02 [24.44, 49.78] \\
        & FTCN & 76.05 [55.36, 99.81] & 3.16 [2.40, 6.48] & 6.14 [5.71, 14.14] & 47.06 [28.57, 61.22] \\
        & TALL & 45.75 [40.54, 65.51] & 3.50 [2.06, 7.14] & 5.19 [4.23, 9.21] & 22.22 [15.71, 33.88] \\
        & FA & 28.54 [25.73, 40.66] & 1.43 [0.96, 5.77] & 5.54 [4.19, 9.69] & 5.84 [4.57, 11.56] \\
        \bottomrule
    \end{tabular*}
\end{table*}

\begin{table*}[t]
    \centering
    \caption{Fairness estimates and 95\% bootstrap confidence intervals
    across the 48 intersectional subgroups defined by race, gender,
    and age. Each entry reports the point estimate followed by its
    confidence interval. Values are percentage-scaled; lower is better.}
    \label{tab:fairness_48_bootstrap_ci}
    \scriptsize
    \setlength{\tabcolsep}{4pt}
    \renewcommand{\arraystretch}{1.05}
    \begin{tabular*}{\textwidth}{@{\extracolsep{\fill}}llcccc@{}}
        \toprule
        \textbf{Training} & \textbf{Detector}
        & $F_{\text{FPR}}$ (CI)
        & $F_{\text{OAE}}$ (CI)
        & $F_{\text{DP}}$ (CI)
        & $F_{\text{MEO}}$ (CI) \\
        \midrule
        \multirow{4}{*}{Balanced (4,743)}
        & Xception & 349.76 [170.80, 458.82] & 10.45 [4.76, 17.19] & 42.86 [30.00, 61.91] & 100.00 [42.86, 100.00] \\
        & FTCN & 604.98 [454.55, 806.17] & 10.45 [7.25, 18.18] & 39.29 [27.58, 58.06] & 100.00 [50.00, 100.00] \\
        & TALL & 164.76 [66.60, 265.47] & 4.48 [2.98, 11.11] & 42.86 [30.00, 62.50] & 33.33 [16.64, 77.83] \\
        & FA & 84.06 [18.58, 162.00] & 1.59 [1.33, 5.89] & 42.86 [29.41, 61.91] & 9.09 [5.56, 30.77] \\
        \midrule
        \multirow{4}{*}{Imbalanced1 (945)}
        & Xception & 397.49 [245.84, 552.57] & 7.46 [4.41, 14.49] & 42.86 [29.23, 61.91] & 100.00 [36.36, 100.00] \\
        & FTCN & 744.14 [535.32, 905.87] & 13.43 [10.53, 22.95] & 42.86 [30.43, 61.91] & 100.00 [72.69, 100.00] \\
        & TALL & 292.86 [183.65, 436.46] & 14.81 [8.57, 24.29] & 42.86 [34.48, 62.50] & 63.64 [37.50, 100.00] \\
        & FA & 561.63 [402.73, 762.97] & 9.30 [8.06, 21.22] & 35.71 [27.12, 54.29] & 100.00 [50.00, 100.00] \\
        \midrule
        \multirow{4}{*}{Imbalanced2 (1,150)}
        & Xception & 555.42 [450.19, 771.18] & 14.93 [9.37, 23.73] & 39.29 [29.51, 58.63] & 90.91 [75.00, 100.00] \\
        & FTCN & 634.55 [444.05, 831.54] & 17.11 [8.62, 28.23] & 43.59 [26.72, 65.68] & 100.00 [83.33, 100.00] \\
        & TALL & 243.46 [147.98, 358.67] & 9.26 [7.84, 19.05] & 46.43 [34.00, 66.67] & 45.45 [25.00, 85.71] \\
        & FA & 116.77 [35.39, 214.53] & 4.17 [2.67, 10.35] & 42.86 [30.00, 61.91] & 20.00 [9.09, 60.00] \\
        \midrule
        \multirow{4}{*}{Imbalanced3 (1,489)}
        & Xception & 162.84 [71.96, 277.22] & 2.99 [2.86, 9.43] & 42.86 [29.63, 61.91] & 20.00 [14.29, 71.43] \\
        & FTCN & 614.25 [457.86, 843.76] & 11.94 [8.11, 20.59] & 42.86 [28.57, 61.91] & 100.00 [66.67, 100.00] \\
        & TALL & 282.68 [177.52, 405.77] & 11.11 [7.81, 20.37] & 42.86 [33.33, 62.50] & 63.64 [33.33, 100.00] \\
        & FA & 84.62 [17.08, 179.02] & 4.92 [3.03, 11.54] & 42.86 [30.00, 61.91] & 20.00 [6.67, 60.00] \\
        \midrule
        \multirow{4}{*}{Imbalanced4 (4,743)}
        & Xception & 335.50 [137.83, 456.47] & 7.46 [3.64, 13.75] & 42.86 [29.49, 61.91] & 100.00 [30.77, 100.00] \\
        & FTCN & 663.38 [481.76, 844.59] & 14.93 [8.96, 23.73] & 42.86 [29.17, 61.91] & 100.00 [83.33, 100.00] \\
        & TALL & 162.34 [65.41, 266.11] & 17.65 [13.84, 29.31] & 42.86 [35.59, 62.50] & 33.33 [20.00, 75.07] \\
        & FA & 86.48 [19.52, 164.15] & 1.69 [1.39, 6.35] & 42.86 [29.41, 61.91] & 9.09 [6.25, 33.33] \\
        \bottomrule
    \end{tabular*}
\end{table*}


\subsection{Shared-Threshold Analysis}
\label{sec:ftcn_shared_threshold}
We evaluate FTCN at one global threshold per operating point, selected using all 291 real videos and then applied unchanged to 290 fake identities and all demographic subgroups. Each fake identity contributes one output from each of the eight generators. Thus, within each generator, the overall TPR is the identity-count-weighted average of subgroup TPRs. We report macro-TPR averaged over the eight generators. Bootstrap resampling is performed at the identity level, retaining all eight generator outputs for each sampled identity.

\begin{table}[htbp]
\centering
\caption{FTCN subgroup macro-TPR at shared operating thresholds. Values are proportions with 95\% identity-clustered bootstrap intervals; $n$ counts fake identities.}
\label{tab:ftcn_subgroups_shared}
\resizebox{\linewidth}{!}{%
\begin{tabular}{llrrr}
\toprule
Attribute & Subgroup & $n$ & T0.1 macro-TPR & T1 macro-TPR \\
\midrule
Gender & Female & 144 & 0.312 [0.259, 0.366] & 0.893 [0.860, 0.924] \\
Gender & Male & 146 & 0.428 [0.367, 0.489] & 0.894 [0.857, 0.927] \\
\midrule
Race & Asian & 71 & 0.338 [0.261, 0.417] & 0.887 [0.833, 0.935] \\
Race & Black & 72 & 0.451 [0.368, 0.537] & 0.915 [0.865, 0.957] \\
Race & Indian & 74 & 0.367 [0.287, 0.448] & 0.904 [0.855, 0.944] \\
Race & White & 73 & 0.325 [0.247, 0.408] & 0.868 [0.820, 0.911] \\
\midrule
Age & 0--9 & 48 & 0.354 [0.258, 0.456] & 0.870 [0.815, 0.919] \\
Age & 10--19 & 48 & 0.375 [0.273, 0.479] & 0.870 [0.794, 0.935] \\
Age & 20--29 & 48 & 0.305 [0.214, 0.401] & 0.893 [0.823, 0.953] \\
Age & 30--39 & 49 & 0.367 [0.270, 0.467] & 0.906 [0.844, 0.957] \\
Age & 40--49 & 49 & 0.429 [0.329, 0.531] & 0.916 [0.860, 0.964] \\
Age & 50+ & 48 & 0.391 [0.287, 0.497] & 0.906 [0.859, 0.948] \\
\bottomrule
\end{tabular}}
\end{table}

Table~\ref{tab:ftcn_subgroups_shared} shows that demographic differences depend on the operating point. The male-minus-female macro-TPR difference is 0.117 at T0.1 (95\% interval: 0.036--0.197), compared with 0.001 at T1 (95\% interval: $-0.047$--0.047). We further test generator-by-demographic interaction using a binomial generalized estimating equation (GEE), with identity as the clustering unit, and an identity-level permutation analysis.
\begin{table}[t]
    \centering
    \caption{FTCN generator-by-demographic interaction tests at
    shared thresholds. Reported $p$-values are nominal.}
    \label{tab:ftcn_interactions}
    \scriptsize
    \setlength{\tabcolsep}{2pt}
    \renewcommand{\arraystretch}{1.05}
    \begin{tabular*}{\linewidth}{@{\extracolsep{\fill}}llrrrr@{}}
        \toprule
        \textbf{Point} & \textbf{Interaction}
        & \shortstack{\textbf{Wald}\\$\chi^2$}
        & \textbf{df}
        & \shortstack{\textbf{GEE}\\$p$}
        & \shortstack{\textbf{Permutation}\\$p$} \\
        \midrule
        \multirow{3}{*}{T0.1}
        & Generator $\times$ Gender & 13.33 & 7  & 0.064 & 0.080 \\
        & Generator $\times$ Race   & 18.86 & 21 & 0.594 & 0.844 \\
        & Generator $\times$ Age    & 57.69 & 35 & 0.009 & 0.0004 \\
        \midrule
        \multirow{3}{*}{T1}
        & Generator $\times$ Gender & 2.80  & 7  & 0.903 & 0.967 \\
        & Generator $\times$ Race   & 25.56 & 21 & 0.224 & 0.199 \\
        & Generator $\times$ Age    & 50.48 & 35 & 0.044 & 0.0024 \\
        \bottomrule
    \end{tabular*}
\end{table}

At the nominal 0.05 level, the age interaction is detected at both operating points, whereas gender and race interactions are not detected. This identifies age-conditioned performance profiles that vary across generators in the evaluated FTCN setting. These performance interactions are distinct from a causal claim about demographic shortcut use.

\section{Supplementary Identity Consistency Evaluation}

\subsection{Controlled Comparison of Identity Pairing}
\label{sec:identity_pairing_control}
We agree that a controlled comparison is necessary to distinguish the effect of identity pairing from other distribution factors. To construct paired and unpaired testing settings with consistent generation conditions, we create two evaluation scenarios:
\begin{itemize}
    \item \textbf{Unpaired (low identity similarity):}
    Real videos from HDTF + fake videos generated from FFHQ identities and HDTF audio. Different identities, same demographic distribution / audio source / generator set / preprocessing.

    \item \textbf{Paired (high identity similarity):}
    Protocol 4 (Subset B), real videos from HDTF + fake videos generated from FFHQ identities and HDTF audio, where real and fake videos share the same identity / demographic distribution/audio source/generator set/preprocessing.
\end{itemize}

We train detectors using the identity-unpaired training set and evaluate them on both the unpaired and paired testing settings. The results show a clear performance gap: when real and fake samples have different identities (Unpaired), all four detectors achieve near-saturated detection performance (AUC $\geq$ 97.97\%). However, when identities are strictly aligned (Paired), all detectors consistently degrade to near-random performance (AUC: 42--55\%, and TPR@FPR=0.1\%: 0--2\%). This indicates that the performance collapse is not simply caused by changes in data source, generation method, background, or image quality, but is strongly related to the removal of identity-based shortcuts.
\begin{table}[t]
    \centering
    \caption{Controlled comparison of identity-unpaired and identity-paired testing settings. All metrics are percentages (higher is better).}
    \label{tab:identity_pairing_control}
    \scriptsize
    \setlength{\tabcolsep}{1.5pt}
    \renewcommand{\arraystretch}{1.05}
    \begin{tabular*}{\linewidth}{@{\extracolsep{\fill}}lllrrrr@{}}
        \toprule
        \multirow{2}{*}{\shortstack{Training\\set}}
        & \multirow{2}{*}{\shortstack{Testing\\set}}
        & \multirow{2}{*}{Detector}
        & \multicolumn{4}{c}{Metrics} \\
        \cmidrule(lr){4-7}
        & & & AUC$\uparrow$ & T1$\uparrow$
        & T0.1$\uparrow$ & ACC$\uparrow$ \\
        \midrule
        \multirow{8}{*}{\shortstack{FairTalking-\\Bench\\(Balanced)}}
        & \multirow{4}{*}{Unpaired}
        & Xception & 99.95 & 98.63 & 92.19 & 99.79 \\
        & & FTCN & 97.97 & 37.03 & 37.03 & 98.86 \\
        & & TALL & 99.91 & 98.53 & 98.53 & 99.33 \\
        & & Adapter & 98.54 & 93.66 & 88.79 & 99.42 \\
        \cmidrule(lr){2-7}
        & \multirow{4}{*}{\shortstack{Paired\\(Subset B)}}
        & Xception & 46.24 & 1.89 & 1.89 & 57.44 \\
        & & FTCN & 42.20 & 0.47 & 0.47 & 42.98 \\
        & & TALL & 46.44 & 0.00 & 0.00 & 27.31 \\
        & & Adapter & 54.54 & 1.16 & 0.26 & 24.69 \\
        \bottomrule
    \end{tabular*}
\end{table}

\section{Additional Robustness Evaluation}

\subsection{Random-Seed Variability}
\label{sec:seed_variability}

Diffusion-based generators introduce stochasticity, and detector performance may be affected by different sampling trajectories. In our initial generation process, we sampled three different random seeds for diffusion-based generators and compared the generated videos using FVD and SyncNet metrics. The results showed only minor variations across different seeds, indicating stable generation quality.

Furthermore, to examine whether the observed generalization patterns (\emph{e.g.}, flow-matching vulnerability) depend on a specific random seed, we generate the training data using different seeds for each diffusion-based generator. As shown in Table~\ref{tab:seed_variability}, we train detectors on diffusion-based training data generated with different random seeds and evaluate them on fake videos generated by flow-matching models. The results remain consistent across different seeds, indicating that the observed vulnerability to flow-matching generators is not caused by a particular sampling trajectory, but reflects the intrinsic distribution gap between diffusion-based and flow-matching generation paradigms.

\begin{table}[t]
    \centering
    \caption{Generation quality and detection performance across random seeds.}
    \label{tab:seed_variability}
    \scriptsize
    \setlength{\tabcolsep}{2pt}
    \renewcommand{\arraystretch}{1.05}
    \begin{tabular*}{\linewidth}{@{\extracolsep{\fill}}lrrrrr@{}}
        \toprule
        \multirow{2}{*}{Generator}
        & \multirow{2}{*}{Seed}
        & \multirow{2}{*}{FVD}
        & \multirow{2}{*}{FVD std}
        & \multicolumn{2}{c}{Flow-Matching} \\
        \cmidrule(lr){5-6}
        & & & & T1 & T0.1 \\
        \midrule
        \multirow{3}{*}{AniPortrait}
        & 42   & 0.0240 & \multirow{3}{*}{0.00048} & 0.24 & 0.01 \\
        & 1024 & 0.0229 &                         & 0.25 & 0.02 \\
        & 3407 & 0.0238 &                         & 0.24 & 0.01 \\
        \midrule
        \multirow{3}{*}{JoyVASA}
        & 42   & 0.0256 & \multirow{3}{*}{0.00079} & 0.13 & 0.02 \\
        & 1024 & 0.0260 &                         & 0.10 & 0.01 \\
        & 3407 & 0.0275 &                         & 0.13 & 0.02 \\
        \midrule
        \multirow{3}{*}{Hallo}
        & 42   & 0.0273 & \multirow{3}{*}{0.00029} & 0.06 & 0.01 \\
        & 1024 & 0.0268 &                         & 0.05 & 0.01 \\
        & 3407 & 0.0275 &                         & 0.05 & 0.01 \\
        \midrule
        \multirow{3}{*}{Ditto}
        & 42   & 0.0262 & \multirow{3}{*}{0.00078} & 0.25 & 0.02 \\
        & 1024 & 0.0246 &                         & 0.26 & 0.02 \\
        & 3407 & 0.0263 &                         & 0.24 & 0.02 \\
        \midrule
        \multirow{3}{*}{Sonic}
        & 42   & 0.0273 & \multirow{3}{*}{0.00073} & 0.15 & 0.02 \\
        & 1024 & 0.0282 &                         & 0.13 & 0.01 \\
        & 3407 & 0.0264 &                         & 0.16 & 0.02 \\
        \bottomrule
    \end{tabular*}
\end{table}

\subsection{Multilingual Evaluation}
\label{sec:multilingual_supp}
As a preliminary study, following our FairTalking-Bench English testing set, we construct Chinese and Spanish testing subsets to evaluate the cross-lingual robustness of the detectors. Specifically, we select 290 audio samples for each language and use the corresponding reference images consistent with the English testing set. They are fed into eight talking-head generators, resulting in 4,640 generated videos (2,320 fake videos per language). We train detectors using our FairTalking-Bench English-based training set and evaluate them on the multilingual testing subsets across English, Chinese, and Spanish, respectively.

As shown in Table~\ref{tab:multilingual_auc}, all four detectors show performance declines on the Chinese test set (Xception: $-3.43\%$, FTCN: $-18.51\%$, TALL: $-2.95\%$, Adapter: $-4.29\%$) and the Spanish test set ($-9.66\%$, $-13.33\%$, $-10.45\%$, and $-12.02\%$, respectively). These results highlight the importance of multilingual evaluation when assessing real-world robustness.

We argue that different languages introduce variations in phonetic patterns, speaking styles, and audio-visual synchronization, which may affect the learned representations of talking-head forgery detectors.

\begin{table}[htbp]
\centering
\caption{AUC across language settings. Parentheses show the change from the English baseline in percentage points, recalculated from the displayed AUCs.}
\label{tab:multilingual_auc}
\begin{tabular}{lrrr}
\toprule
Detector & English & Chinese & Spanish \\
\midrule
Xception & 99.98 & 96.55 ($-3.43$) & 86.89 ($-13.09$) \\
FTCN & 99.54 & 86.21 ($-13.33$) & 81.03 ($-18.51$) \\
TALL & 99.92 & 96.97 ($-2.95$) & 89.47 ($-10.45$) \\
Adapter & 99.99 & 95.70 ($-4.29$) & 87.97 ($-12.02$) \\
\bottomrule
\end{tabular}
\end{table}


\section{User Study Instructions}
\label{sec:userstudy_instructions}

\paragraph{Muted Visual Study.}
Participants were presented with 24 sets of videos randomly sampled from four datasets (FF++, CelebDF++, THB, and FairTalking-Bench). All videos were muted to ensure fair comparison, as FF++ and CelebDF++ lack audio tracks. Participants were asked to rate each video on a 1--4 scale:
\begin{itemize}[leftmargin=*]
      \item Score 1: Clearly fake, obvious artifacts visible.
      \item Score 2: Likely fake, some noticeable artifacts.
      \item Score 3: Likely real, minor imperfections.
      \item Score 4: Highly realistic, indistinguishable from real.
\end{itemize}

\paragraph{Audiovisual Study.}
Participants were shown 12 paired video sets, each containing one video from our dataset and one from THB~\cite{Xiong_2026_WACV}, both with audio enabled. For each pair, participants answered two questions:
\begin{itemize}[leftmargin=*]
      \item ``Which video has more realistic visual appearance (natural expressions and head movements)?''
      \item ``Which video has better lip-audio synchronization?''
\end{itemize}
The presentation order of videos within each pair was randomized to avoid ordering bias.

\section{Limitations}
\label{sec:limitations}

While FairTalking-Bench represents a significant step toward fair and comprehensive deepfake detection evaluation, we acknowledge several limitations:

\begin{itemize}[leftmargin=*]

\item \textbf{Demographic Taxonomy Granularity.} We adopt the FairFace taxonomy with 4 racial groups, 2 genders, and 6 age intervals. This categorization, while consistent with community standards, does not
  capture finer-grained distinctions (\emph{e.g.}, mixed-race individuals, non-binary gender identities, or ethnicity-level differences). Future work could explore more inclusive and fine-grained demographic
  annotations.

\item \textbf{Scope of Identity-Paired Validation.} The independent evaluation includes 126 participants, each contributing three real videos. Applying eight generators yields 378 real and 3,024 fake videos, for a total of 3,402 videos. This extends the evaluation beyond HDTF; broader capture environments and participant populations remain directions for further evaluation.

\item \textbf{Language and Audio Diversity.} The reference audio clips are sourced from English-language TED talks (LIUM dataset). The supplementary Chinese and Spanish evaluation (Section~\ref{sec:multilingual_supp}) extends the tested language settings, but the primary collection remains English-based and does not cover the full range of languages, accents, and recording conditions.

\item \textbf{Computational Cost.} Generating 23,712 fake videos using 8 different generators requires substantial computational resources. We acknowledge that reproducing the full benchmark or extending it with
  additional generators may pose accessibility challenges for resource-limited research groups.

\end{itemize}
These limitations represent important directions for future work and do not diminish the core contributions of our benchmark in revealing critical fairness, generalization, and identity-paired evaluation challenges in current deepfake detection systems.

\section{Broader Impact}
\label{sec:broader_impact}
                            
\textbf{Positive Societal Impact.} Our work contributes to the development of fairer and more robust deepfake detection systems. By providing a demographically balanced benchmark, FairTalking-Bench enables researchers to identify and mitigate biases in detection algorithms that could otherwise disproportionately affect underrepresented demographic groups. This is particularly important as deepfake detection technologies are increasingly deployed in real-world scenarios such as content moderation, identity verification, and combating disinformation campaigns. Furthermore, our systematic evaluation protocols reveal critical vulnerabilities in existing detectors (e.g., low performance under identity-paired synthesis and unseen generators), which can guide the community toward developing more reliable and equitable detection solutions.                      
 
\textbf{Potential Negative Societal Impact.} We acknowledge two primary concerns: (1) The dataset contains high-fidelity synthesized deepfake videos generated by state-of-the-art talking-head methods. If misused,
these videos or the generation pipeline could serve as resources for creating more convincing forgeries for malicious purposes such as disinformation, fraud, or non-consensual impersonation. (2) Our detailed analysis of detector weaknesses—particularly the performance reduction under identity-paired synthesis and on challenging unseen generators—could potentially be exploited by adversaries seeking to craft deepfakes that evade current detection systems.                                                                                     
\textbf{Mitigation Strategies.} To minimize potential harms, we adopt the following measures:
\begin{itemize}[leftmargin=*]
\item \textbf{Restricted License:} The dataset will be released under a CC BY-NC 4.0 license, restricting usage to non-commercial academic research purposes only.
\item \textbf{Access Control:} Users are required to sign a data usage agreement explicitly prohibiting any malicious applications, redistribution, or use for generating deceptive content.
\item \textbf{Data Sources:} All real videos are derived from publicly available datasets (DFDC, CelebV-HQ, HDTF) with adherence to their original licensing and access policies. Reference images follow the FFHQ license.
\item \textbf{Identity Separation:} The standard benchmark uses disjoint real and fake identity pools and mutually exclusive identities across splits. The independently collected identity-paired evaluation is described separately in Protocol~4.
\item \textbf{Defense-Oriented:} Our work is fundamentally defensive in nature—the benchmark is designed to strengthen detection capabilities rather than to advance generation quality.
\end{itemize}
We believe that the societal benefits of enabling fair, robust, and generalizable deepfake detection substantially outweigh the potential risks, particularly given the above mitigation strategies.

\section{Future Work}
We plan to extend FairTalking-Bench along five directions:
\begin{enumerate}
    \item More driving modalities, including video and text, beyond audio;
    \item More diverse audio sources, including Mandarin, regional dialects, Spanish, and French;
    \item More advanced generators, including stronger open-source models, closed-source models, and commercial systems;
    \item Increased identity diversity;
    \item More inclusive demographic categories.
\end{enumerate}

We plan to establish a community contribution mechanism with standardized data formats and evaluation protocols, allowing researchers to contribute new generators and evaluation results. We envision FairTalking-Bench as an evolving benchmark that continuously adapts to advances in generative technologies.

\end{document}